\documentclass{article}

\usepackage{hyperref}
\usepackage{natbib}
\usepackage{fullpage}
\usepackage{graphicx}
\usepackage{times}
\usepackage{amsmath,amsfonts,amssymb,latexsym,graphicx}
\usepackage{enumerate}
\usepackage[linesnumbered,boxed,ruled,vlined]{algorithm2e}
\usepackage{tabularx}

\usepackage{delarray}
\usepackage{times}
\usepackage{subfigure}
\usepackage{color}
\usepackage{amssymb}
\usepackage{amsmath}
\usepackage{tikz}
\usepackage{xspace}
\usepackage{enumerate}
\usepackage{multirow}
\usepackage{nameref}
\usepackage{booktabs}
\usepackage{mdframed}
\usepackage{fmtcount}
\usepackage{footnote}

\newtheorem{theorem}{Theorem}
\newtheorem{lemma}{Lemma}
\newtheorem{corollary}{Corollary}

\newtheorem{assumption}{Assumption}

\hypersetup{
    colorlinks=true,       % false: boxed links; true: colored links
    linkcolor=black,    %cyan,        % color of internal links
    citecolor=blue,        % color of links to bibliography
    filecolor=magenta,     % color of file links
    urlcolor=blue
}
\newcommand*{\qed}{\hfill\ensuremath{\square}}

\newcommand{\eat}[1]{}

\newcommand{\E}{\mathbb{E}}
\newcommand{\D}{\mathrm{d}}
\newcommand{\T}{^\intercal}
\newcommand{\Gen}{\mathcal{L}}
\newcommand{\N}{\mathrm{N}}
\newcommand{\rhmc}{\mathsf{R^{HMC}}}
\newcommand{\rld}{\mathsf{R^{LD}}}
\newcommand{\re}[1]{\textcolor{black}{#1}}

\begin{document}

\title{Stochastic Gradient Hamiltonian Monte Carlo with Variance Reduction for Bayesian Inference}

\author{ Zhize Li\footnotemark[2] \qquad
Tianyi Zhang\footnotemark[2] \qquad
Shuyu Cheng \qquad
Jun Zhu \qquad
Jian Li \\
     Tsinghua University\\
\{zz-li14, tianyi-z16, chengsy18\}@mails.tsinghua.edu.cn\\
\{dcszj, lijian83\}@mail.tsinghua.edu.cn
}
\renewcommand{\thefootnote}{\fnsymbol{footnote}}
\footnotetext[2]{denotes equal contribution}

\date{}
\maketitle

\begin{abstract}
Gradient-based Monte Carlo sampling algorithms, like Langevin dynamics and Hamiltonian Monte Carlo, are important methods for Bayesian inference.
In large-scale settings, full-gradients are not affordable and thus stochastic gradients evaluated on mini-batches are used as a replacement.
In order to reduce the high variance of noisy stochastic gradients, \cite{langevin} applied the standard variance reduction technique on stochastic gradient Langevin dynamics and obtained both theoretical and experimental improvements.
In this paper, we apply the variance reduction tricks on Hamiltonian Monte Carlo and achieve better theoretical convergence results compared with the variance-reduced Langevin dynamics.
Moreover, we apply the symmetric splitting scheme in our variance-reduced Hamiltonian Monte Carlo algorithms to further improve the theoretical results.
The experimental results are also consistent with the theoretical results.
As our experiment shows, variance-reduced Hamiltonian Monte Carlo demonstrates better performance than variance-reduced Langevin dynamics in Bayesian regression and classification tasks on real-world datasets.
\end{abstract}

\section{Introduction}
Gradient-based Monte Carlo algorithms are useful tools for sampling posterior distributions. Similar to gradient descent algorithms, gradient-based Monte Carlo generates posterior samples iteratively using the gradient of log-likelihood.

Langevin dynamics (LD) and Hamiltonian Monte Carlo (HMC) \citep{duane1987hybrid,neal2011mcmc} are two important examples of gradient-based Monte Carlo sampling algorithms that are widely used in Bayesian inference. Since calculating likelihood on large datasets is expensive, people use stochastic gradients \citep{robbins1951stochastic} in place of  full gradient, and have, for both Langevin dynamics and Hamiltonian Monte Carlo, developed their stochastic gradient counterparts \citep{welling2011bayesian,chen2014stochastic}. Stochastic gradient Hamiltonian Monte Carlo (SGHMC) usually converges faster than stochastic gradient Langevin dynamics (SGLD) in practical machine learning tasks like covariance estimation of bivariate Gaussian and Bayesian neural networks for classification on MNIST dataset, as demonstrated in \citep{welling2011bayesian}. Similar phenomenon was also observed in \citep{chen2015convergence} where SGHMC and SGLD were compared on both synthetic and real-world datasets. Intuitively speaking, comparing against SGLD, SGHMC has a momentum term that may enable it to explore the parameter space of posterior distribution much faster when the gradient of log-likelihood becomes smaller.

Very recently, \cite{langevin} borrowed the standard variance reduction techniques from the stochastic optimization literature \citep{johnson2013accelerating,defazio2014saga} and applied them on SGLD to obtain two variance-reduced SGLD algorithms (called SAGA-LD and SVRG-LD) with improved theoretical results and practical performance. Because of the superiority of SGHMC over SGLD in terms of convergence rate in a wide range of machine learning tasks, it would be a natural question whether such variance reduction techniques can be applied on SGHMC to achieve better results than variance-reduced SGLD.

\re{The challenge is that SGHMC is more complicated than SGLD, i.e., the extra momentum term (try to explore faster) and friction term (control the noise caused by SGHMC from HMC) in SGHMC. Note that the friction term in SGHMC is inherently different than SGLD since LD itself already has noise so it can be directly extended to SGLD, while HMC itself is deterministic.
To the best of our knowledge, there is even no existing work to prove that SGHMC is better than SGLD.
So in this paper we need to give some new approaches and insights in our analysis to prove that variance-reduced SGHMC is better than variance-reduced SGLD due to the existence of momentum term and friction term.
}
Note that in stochastic optimization literature, the variance-reduced methods with momentum term (e.g., \citep{allen2017katyusha,lan2019unified}) indeed are better than variance-reduced methods without momentum term (e.g., \citep{johnson2013accelerating}) especially for convex optimization.

Actually, it seems that the variance reduction in this stochastic Bayesian inference is more effective compared with stochastic optimization settings.
Intuitively, the full gradient case (no variance) may converge to a saddle point or a local minimum (not a global minimum) in nonconvex optimization, and the variance of the stochastic gradient estimator may be useful for escaping saddle points or bad local minima. Thus, we may not want to reduce the variance.
However, the full gradient case (no variance) will converge to the stationary posterior distribution for Bayesian inference. Thus, it is useful to reduce the variance of the stochastic gradient estimator for obtaining more approximate posterior distribution.
Note that in large-scale settings, full-gradients (no variance) are not affordable and thus stochastic gradients evaluated on mini-batches are used as a replacement.

\subsection{Our contribution}
\begin{enumerate}
  \item We propose two variance-reduced versions of Hamiltonian Monte Carlo algorithms (called SVRG-HMC and SAGA-HMC) using the standard approaches from \citep{johnson2013accelerating,defazio2014saga}.
      Compared with SVRG/SAGA-LD \citep{langevin}, our algorithms guarantee improved theoretical convergence results due to the extra momentum term in HMC (see Corollary \ref{cor:comp}).
      %Concretely, the proposed SVRG/SAGA-HMC are $O(\frac{n^2}{bL})$ times faster than SVRG/SAGA-LD if $D\ge{1}/{L\sqrt{h}}$, in terms of the convergence bound related to the variance reduction.
      %Otherwise, SVRG/SAGA-HMC are $O(D^2)$ times faster than SVRG/SAGA-LD.
  \item Moreover, we combine the proposed SVRG/SAGA-HMC algorithms with the symmetric splitting scheme \citep{chen2015convergence,leimkuhler2016adaptive} to extend them to 2nd-order integrators, which further improve the dependency on step size (see the difference between Theorem \ref{vrhmc-svrg} and \ref{svrg2nd}).
      We denote these two algorithms as SVRG2nd-HMC and SAGA2nd-HMC.
  \item Finally, we evaluate our algorithms on real-world datasets and compare them with SVRG/SAGA-LD \citep{langevin}; as it turns out, our algorithms converge markedly faster than the benchmarks (vanilla SGHMC and SVRG/SAGA-LD).
\end{enumerate}

\subsection{Related work}

Langevin dynamics and Hamiltonian Monte Carlo are two important sampling algorithms that are widely used in Bayesian inference.
Many literatures studied how to develop the variants of them to achieve improved performance, especially for scalability for large datasets.
\cite{welling2011bayesian} started this direction with the notable work stochastic gradient Langevin dynamics (SGLD).
\cite{ahn2012bayesian} proposed a modification to SGLD reminiscent of Fisher scoring to better estimate the gradient noise variance, with lower classification error rates on HHP dataset and MNIST dataset.
\cite{chen2014stochastic} developed the stochastic gradient version of HMC (SGHMC), with a quite nontrivial approach different from SGLD.
\cite{ding2014bayesian} further improved SGHMC by a new dynamics to better control the gradient noise, and the proposed stochastic gradient Nos\'{e}-Hoover thermostats (SGNHT) outperforms SGHMC on MNIST dataset.

Various settings of Markov Chain Monte Carlo (MCMC) are also considered.
\cite{girolami2011riemann} enhanced LD and HMC by exploring the Riemannian structure of the target distribution, with Riemannian manifold LD and HMC (RMLD and RMHMC, respectively).
\cite{byrne2013geodesic} developed geodesic Monte Carlo (GMC) that is applicable to Riemannian manifolds with no global coordinate systems.
Large-scale variants of RMLD, RMHMC and GMC with stochastic gradient were developed by \citep{patterson2013stochastic}, \citep{ma2015complete} and \citep{liu2016stochastic}, respectively.
\cite{ahn2014distributed} studied the behaviour of stochastic gradient MCMC algorithms for distributed posterior inference.
\re{Very recently, \cite{zou2018stochastic} used a stochastic variance-reduced HMC for sampling from smooth and strongly log-concave distributions which requires $f$ is smooth and strongly convex. In this paper, we do not assume $f$ is strongly convex or convex and we also use an efficient discretization scheme to further imporve the convergence results. Their results were measured with 2-Wasserstein distance, while ours are measured with mean square error.}
Note that the variance reduction techniques have already been used in nonconvex optimization literature (see e.g., \citep{allen2016variance,reddi2016stochastic,li2018simple,ge2019stable,li2019ssrgd}),
and they achieved improved convergence results.

\section{Preliminary}

Let $X = \{x_i\}_{i=1}^n$ be a $d$-dimensional dataset that follows the distribution $\Pr(X|\theta) = \prod_{i=1}^n \Pr(x_i|\theta)$.
Then, we are interested in sampling the posterior distribution $\Pr(\theta|X)\propto \Pr(\theta)\prod_{i=1}^n \Pr(x_i|\theta)$ based on Hamiltonian Monte Carlo algorithms.
Let $[n]$ denote the set $\{1, 2, \ldots, n\}$. Define $f(\theta) = \sum_{i=1}^n f_i(\theta)-\log\Pr(\theta)$, where $f_i(\theta) = -\log\Pr(x_i|\theta)$ and $i \in [n]$. Similar to \citep{langevin}, we assume that each $f_i$ is $L$-smooth and $G$-Lipschitz, for all $i\in [n]$.

The  general algorithmic framework maintains two sequences for $t = 0, 1, \ldots, T-1$ by the following discrete time procedure:
\begin{align}
p_{t+1} &= (1-Dh)p_t -h\tilde{\nabla}_t + \sqrt{2Dh}\cdot\xi_t \label{eq:d1}\\
\theta_{t+1} &= \theta_t + hp_{t+1}
\end{align}
and then returns the samples $\{\theta_1, \theta_2, \ldots, \theta_T\}$ as an approximation to the stationary distribution $\Pr(\theta|X)$. $\theta_t$ is the parameter we wish to sample and $p_t$ is an auxiliary variable conventionally called the ``momentum''. Here $h$ is step size, $D$ is a constant independent of $\theta$ and $p$, $\xi_t\sim \mathrm{N}(0, I_d)$ and $\tilde{\nabla}_t$ is a mini-batch approximation of the full gradient $\nabla f(\theta_t)$.
If we set $\tilde{\nabla}_t = \frac{n}{b}\sum_{i\in I} \nabla f_i(\theta_t)$, $I$ being a $b$-element index set uniformly randomly drawn (with replacement) from $\{1, 2, \ldots, n\}$ as introduced in \citep{robbins1951stochastic}, then the algorithm becomes SGHMC.

The above discrete time procedure provides an approximation to the continuous Hamiltonian Monte Carlo diffusion process $(\theta, p)$:
\begin{align}
\D \theta  &= p\D t\\
\D p&= -\nabla_\theta f(\theta) \D t - Dp \D t+ \sqrt{2D} \D W
\end{align}
Here $W$ is a Wiener process. According to \citep{chen2015convergence}, the stationary joint distribution of $(\theta, p)$ is $\pi(\theta,p) \propto e^{-f(\theta) - \frac{p\T p}{2}}$.

How do we evaluate the quality of the samples $\{\theta_1, \theta_2, \ldots, \theta_T\}$?  Assuming $\phi: \mathbb{R}^d\rightarrow \mathbb{R}$ is a smooth test function, we wish to upper bound the Mean-Squared Error (MSE) $\E(\hat{\phi} - \bar{\phi})^2$, where $\hat{\phi} = \frac{1}{T}\sum_{t=1}^T \phi(\theta_t)$ is the empirical average, and $\bar{\phi} = \E_{\theta\sim \Pr(\theta \mid X)}\phi(\theta)$ is the population average. So, the objective of our algorithm is to carefully design $\tilde{\nabla}_t$ to minimize $\E(\hat{\phi} - \bar{\phi})^2$ in a faster way, where $\tilde{\nabla}_t$ is a stochastic approximation of $\nabla f(\theta_t)$.

To study how the choice of $\tilde{\nabla}_t$ influences the value of $\E(\hat{\phi} - \bar{\phi})^2$, define $\psi(\theta, p)$ to be the solution to the Poisson equation $\Gen \psi = \phi(\theta) - \bar{\phi}$, $\Gen$ being the generator of Hamiltonian Monte Carlo diffusion process. In order to analyze the theoretical convergence results related to the MSE $\E(\hat{\phi} - \bar{\phi})^2$, we inherit the following assumption from \citep{chen2015convergence}.

\begin{assumption}[\citep{chen2015convergence}]\label{3rd}
Function $\psi$ is bounded up to 3rd-order derivatives by some real-valued function $\Gamma(\theta, p)$, i.e. $\|\mathcal{D}^k \psi\|\leq C_k \Gamma^{q_k}$ where $\mathcal{D}^k$ is the $k$th order derivative for $k = 0, 1, 2, 3$, and $C_k, q_k > 0$.
Furthermore, the expectation of $\Gamma$ on $\{(\theta_t, p_t) \}$ is bounded, i.e. $\sup_t \E [\Gamma^q(\theta_t, p_t)] < \infty$ and that $\Gamma$ is smooth such that $\sup_{s\in (0, 1)}\Gamma^q(s\theta + (1-s)\theta^\prime, s p+ (1-s)p^\prime)\leq C(\Gamma^q(\theta, p) + \Gamma^q(\theta^\prime, p^\prime))$, $\forall \theta, p, \theta^\prime, p^\prime, q\leq \max 2q_k$ for some constant $C > 0$.
\end{assumption}

Define operator $\Delta V_t = (\tilde{\nabla}_t - \nabla f(\theta_t))\cdot \nabla$ for all $t = 0, 1, 2, \ldots, T-1$. When the above assumption holds, we have the following theorem by \citep{chen2015convergence}.

For the rest of this paper, for any two values $A, B > 0$, we say $A\lesssim B$ if $A = O(B)$, where the notation $O(\cdot)$ only hides a constant factor independent of algorithm parameters $T, n, D, h, G, b$.

\begin{theorem}[\citep{chen2015convergence}]\label{operator}
Let $\tilde{\nabla}_t$ be an unbiased estimate of $\nabla f(\theta_t)$ for all $t$. Then under assumption \ref{3rd}, for a smooth test function $\phi$, the MSE of SGHMC is bounded in the following way:
\begin{equation}\label{eq:base}
\E(\hat{\phi} - \bar{\phi})^2\lesssim \frac{\frac{1}{T}\sum_{t=0}^{T-1}\E(\Delta V_t \psi(\theta_t, p_t))^2}{T} + \frac{1}{Th} + h^2
\end{equation}
\end{theorem}

Similar to the [A2] assumption in \citep{langevin}, we also need to make the following assumption which relates $\Delta V_t\phi(\theta, p)$ to the difference $\|\tilde{\nabla}_t- \nabla f(\theta)\|^2$.
\begin{assumption}\label{strong}
$(\Delta V_t \psi(\theta_t, p_t))^2\lesssim \|\tilde{\nabla}_t - \nabla f(\theta_t)\|^2$ for all $0\leq t < T$.
\end{assumption}

Combined with Theorem \ref{operator}, Assumption \ref{strong} immediately yields the following corollary.

\begin{corollary}\label{noise}
Under Assumptions \ref{3rd} and \ref{strong}, we have:
\begin{equation*}
\E(\hat{\phi} - \bar{\phi})^2\lesssim \frac{\frac{1}{T}\sum_{t=0}^{T-1}\E\|\tilde{\nabla}_t - \nabla f(\theta_t)\|^2}{T} + \frac{1}{Th} + h^2
\end{equation*}
\end{corollary}

As we mentioned before, if we take $\tilde{\nabla}_t$ to be the Robbins \& Monro approximation of $\nabla f(\theta_t)$ \citep{robbins1951stochastic}, then it becomes SGHMC and the following corollary holds since all $f_i$'s are $G$-Lipschitz and $\E(X - \E X)^2\leq \E X^2$ for any random variable $X$.
\begin{corollary}\label{sghmc}
Under Assumptions \ref{3rd} and \ref{strong}, the MSE of SGHMC is bounded as:
\begin{align}
\E(\hat{\phi} - \bar{\phi})^2 &\lesssim \frac{\frac{1}{T}\sum_{t=0}^{T-1}\E\|\tilde{\nabla}_t - \nabla f(\theta_t)\|^2}{T} + \frac{1}{Th} + h^2 \notag \\
&\leq \frac{n^2 G^2}{bT} + \frac{1}{Th} + h^2 \label{eq:sghmc}
\end{align}
\end{corollary}

\section{Variance Reduction for Hamiltonian Monte Carlo}
In this section, we introduce two versions of variance-reduced Hamiltonian Monte Carlo based on SVRG \citep{johnson2013accelerating} and SAGA \citep{defazio2014saga} respectively.

\subsection{SVRG-HMC}
In this subsection, we propose the SVRG-HMC algorithm (see Algorithm \ref{hmc-svrg}) which is based on the SVRG algorithm. As can be seen from Line 8 of Algorithm \ref{hmc-svrg}, we use $\tilde{\nabla}_{tK + k} = -\nabla\log\Pr(\theta_{tK + k}) + \frac{n}{b}\sum_{i\in I}\big(\nabla f_i(\theta_{tK + k})- \nabla f_i(w)\big) + g$ , where $g$ is $\sum_{i=1}^n \nabla f_i(\theta_{tK})$, as the stochastic estimation for the full gradient $\nabla f(\theta_{tK + k})$.

Note that we initialize $\theta_0, p_0$ to be zero vectors in the algorithm only to simplify the theoretical analysis.
It would still work with an arbitrary initialization.

\begin{algorithm}[!htb]
	\caption{SVRG-HMC}\label{hmc-svrg}
	parameters $T, K, b, h > 0$, $Dh < 1$, $D\geq 1$\;
	initialize $\theta_0 = p_0 = 0$\;
	\For{$t = 0, 1, \ldots, T / K-1$}{
		compute $g = \sum_{i=1}^n \nabla f_i(\theta_{tK})$\;
		$w = \theta_{tK}$\;
		\For{$k = 0, 1, \ldots, K-1$}{
			uniformly sample an index subset $I\subseteq [n]$, $|I| = b$\;
			$\tilde{\nabla}_{tK + k} = -\nabla\log\Pr(\theta_{tK + k}) + \frac{n}{b}\sum_{i\in I}\Big(\nabla f_i(\theta_{tK + k}) - \nabla f_i(w)\Big) + g$\;
			$p_{tK + k+1} = (1-Dh)p_{tK + k} - h\tilde{\nabla}_{tK+k} + \sqrt{2Dh}\xi_{tK + k}$\;
			$\theta_{tK + k+1} = \theta_{tK + k} + hp_{tK + k+1}$\;
		}
	}
	\Return $\{\theta_t\}_{1\leq t\leq T}$\;
\end{algorithm}

The following theorem shows the convergence result for MSE of SVRG-HMC (Algorithm \ref{hmc-svrg}).
We defer all the proofs to Appendix \ref{app:proofs}.
\begin{theorem}\label{vrhmc-svrg}
	Under Assumptions \ref{3rd} and \ref{strong}, the MSE of SVRG-HMC is bounded as:
\begin{equation}	
\mathbb{E}[(\hat{\phi} - \bar{\phi})^2]
	\lesssim \min\Big\{\frac{n^2G^2}{bT}, \frac{L^2 n^2 K^2h^2}{bT}\Big(\frac{\sqrt{n^2G^2 + D^2d}}{D-L^2n^2K^2h^3b^{-1}}\Big)^2\Big\} + \frac{1}{Th} + h^2 \label{eq:svrg-hmc}
\end{equation}
\end{theorem}

To see how the SVRG-HMC (Algorithm \ref{hmc-svrg}) is compared with SVRG-LD \citep{langevin}, we restate their results as following.

\begin{theorem}[\citep{langevin}]\label{benchmark}
	Under Assumptions \ref{3rd} and \ref{strong}, the MSE of SVRG-LD is bounded as:
\begin{equation}
\mathbb{E}[(\hat{\phi} - \bar{\phi})^2] \lesssim \frac{\min\{n^2G^2, n^2K^2(n^2L^2h^2G^2 + hd) \}}{bT} + \frac{1}{Th} + h^2 \label{eq:svrg-ld}
\end{equation}
\end{theorem}

We assume $n^2G^2 > n^2K^2(n^2L^2 h^2G^2 + hd)$. Otherwise the MSE upper bound of SVRG-LD would be equal to SGHMC (see (\ref{eq:sghmc}) and (\ref{eq:svrg-ld})). We then omit the same terms (i.e., second and third terms) in the RHS of (\ref{eq:svrg-hmc}) and (\ref{eq:svrg-ld}). Then we have the following lemma.

\begin{lemma}\label{lm:comp}
Let $\rhmc = \frac{L^2 n^2 K^2h^2}{bT}\big(\frac{\sqrt{n^2G^2 + D^2d}}{D-L^2n^2K^2h^3b^{-1}}\big)^2$ (i.e., the first term in the RHS of (\ref{eq:svrg-hmc}))
and $\rld = \frac{n^2K^2(n^2L^2h^2G^2 + hd)}{bT}$ (i.e., the first term in the RHS of (\ref{eq:svrg-ld})). Then the following inequality holds.
\begin{equation}
 \rhmc\leq \max\Big\{\frac{1}{D^2}, \frac{L}{nK}\Big\}\rld \label{eq:comp}
\end{equation}
In particular, if $D\ge{1}/{L\sqrt{h}}$, then (\ref{eq:comp}) becomes:
\begin{equation}
\rhmc \leq \frac{L}{nK} \rld \label{eq:comp2}
\end{equation}
\end{lemma}

Note that $K$ is suggested to be $2n$ by \citep{johnson2013accelerating} or $n/b$ by \citep{langevin}.
we obtain the following corollary from Lemma \ref{lm:comp}.
\begin{corollary}\label{cor:comp}
If $K$ is $n/b$ as suggested by \citep{langevin}, then (\ref{eq:comp2}) becomes:
\begin{equation}
\rhmc \leq \frac{bL}{n^2} \rld  \label{eq:comp3}
\end{equation}
In other words, the SVRG-HMC is $O(\frac{n^2}{bL})$ times faster than SVRG-LD, in terms of the convergence bound related to the variance reduction (i.e., the first terms in RHS of (\ref{eq:svrg-hmc}) and (\ref{eq:svrg-ld})).
Note that $n$ is the size of dataset which can be very large, b is the mini-batch size which is usually a small constant and $L$ is the Lipschitz smooth parameter for $f_i(\theta)$.
\end{corollary}

We also want to mention that the convergence proof for SVRG-HMC (i.e., Theorem \ref{vrhmc-svrg}) is a bit more difficult than that for SVRG-LD \cite{langevin} due to the momentum variable $p$ (see Line 9 of Algorithm \ref{hmc-svrg}).
Concretely, the main part of the proof in both SVRG-LD and SVRG-HMC is to bound the variance.
Moreover, the variance can be bounded by the adjacent distance $\{\|{\theta_t-\theta_{t-1}}\|^2\}$ in both SVRG-LD and SVRG-HMC.
In SVRG-LD \citep{langevin}, they can directly bound each of $\|{\theta_t-\theta_{t-1}}\|^2$ for $t\in[T]$. However, due to the momentum variable $p$ in our SVRG-HMC (see Line 9 of Algorithm \ref{hmc-svrg}), the distances $\{\|{\theta_t-\theta_{t-1}}\|^2\}_{t\in [T]}$ are  more correlated. Thus, we cannot directly bound each of $\|{\theta_t-\theta_{t-1}}\|^2$ independently like SVRG-LD. We bound the variance as a whole, i.e., we bound the summation of the variance which is equivalent to bound the summation $\sum_{t\in [T]}\|{\theta_t-\theta_{t-1}}\|^2$. Then we get a quadratic inequality due to the correlation among $\{\|{\theta_t-\theta_{t-1}}\|^2\}_{t\in [T]}$.
Finally, we solve this quadratic inequality to bound the variance.

\subsection{SAGA-HMC}
In this subsection, we propose the SAGA-HMC algorithm by applying the SAGA framework \citep{defazio2014saga} to the Hamiltonian Monte Carlo. The details are described in Algorithm \ref{hmc-saga}.
Similar to the SVRG-HMC, we initialize $\theta_0, p_0$ to be zero vectors in the algorithm only to simplify the analysis; it would still work with an arbitrary initialization.

\begin{algorithm}[!h]
	\caption{SAGA-HMC}\label{hmc-saga}
	parameters $T, b, h > 0$, $Dh < 1$, $D\geq 1$\;
	initialize $\theta_0 = 0, p_0 = 0$\;
	initialize an array $\alpha_0^i = \theta_0, \forall i\in [n]$\;
	compute $g = \sum_{i=1}^n \nabla f_i(\alpha_0^i)$\;
	\For{$t = 0, 1, \ldots, T-1$}{
		uniformly randomly pick a set $I\subseteq [n]$ such that $|I| = b$\;
		$\tilde{\nabla}_t = -\nabla\log \Pr(\theta_t) + \frac{n}{b}\sum_{i\in I}(\nabla f_i(\theta_t) - \nabla f_i(\alpha_t^i)) + g$\;
		$p_{t+1} = (1 - Dh)p_t - h\tilde{\nabla}_t + \sqrt{2Dh} \xi_t$\;
		$\theta_{t+1} = \theta_t + h p_{t+1}$\;
		update $\alpha_{t+1}^i = \theta_t$, $\forall i\in I$\;
		$g\leftarrow g + \sum_{i\in I}(\nabla f_i(\alpha_{t+1}^i) - \nabla f_i(\alpha_t^i))$\;
	}	
	\Return $\{\theta_t\}_{t=1}^T$\;
\end{algorithm}

The following theorem shows the convergence result for MSE of SAGA-HMC. The proof is deferred to Appendix \ref{app:proofs}.
\begin{theorem}\label{vrhmc-saga}
	Under Assumptions \ref{3rd} and \ref{strong}, the MSE of SAGA-HMC is bounded as:
	$$\mathbb{E}[(\hat{\phi} - \bar{\phi})^2] \lesssim
\min\Big\{\frac{n^2G^2}{bT}, \frac{L^2 n^4h^2 }{T b^3}\Big(\frac{\sqrt{n^2G^2 + D^2d}}{D-L^2n^4h^3b^{-3}}\Big)^2\Big\} + \frac{1}{Th} + h^2$$
\end{theorem}

Note that SAGA-HMC can be compared with SAGA-LD \citep{langevin} in a very similar manner to SVRG-HMC (i.e., Lemma \ref{lm:comp} and Corollary \ref{cor:comp}). Thus we omit such a repetition.

\section{Variance-reduced SGHMC with Symmetric Splitting}
Symmetric splitting is a numerically efficient method introduced in \citep{leimkuhler2016adaptive} to accelerate the gradient-based algorithms.
We note that one additional advantage of SGHMC over SGLD is that SGHMC can be combined with symmetric splitting while SGLD cannot \citep{chen2015convergence}.
So it is quite natural to combine symmetric splitting with the proposed SVRG-HMC and SAGA-HMC respectively to see if any further improvements can be obtained.

The symmetric splitting scheme breaks the original recursion into 5 steps:
\begin{align}
&\theta_t^{(1)} = \theta_t + \frac{h}{2}p_t	\label{eq:dsym1}\\
&p_t^{(1)} = e^{-Dh/2}p_t\\
& p_t^{(2)} = p_t^{(1)} -h\tilde{\nabla}_t + \sqrt{2Dh}\xi_t \label{eq:dsym3}\\
&p_{t+1} = e^{-Dh/2}p_t^{(2)}\\
& \theta_{t+1} = \theta_t^{(1)} + \frac{h}{2}p_{t+1}
\end{align}
If we eliminate the intermediate variables, then
\begin{align}
p_{t+1} &= e^{-Dh/2} \big(e^{-Dh/2}p_t - h\tilde{\nabla}_t + \sqrt{2Dh}\xi_t\big)\\
\theta_{t+1} &= \theta_t + \frac{h}{2} p_{t+1} + \frac{h}{2}p_t
\end{align}

Same as before $\xi_t\sim \N(0, I_d)$. Note that the stochastic gradient $\tilde{\nabla}_t$ is computed at $\theta_t^{(1)}$ (which is $\theta_t + \frac{h}{2}p_t$) instead of $\theta_t$ (see (\ref{eq:d1}), (\ref{eq:dsym1}) and (\ref{eq:dsym3})).
As shown in \citep{chen2015convergence}, this symmetric splitting scheme is a 2nd-order local integrator.
Then it improves the dependency of MSE on step size $h$, i.e., the third term in the RHS of (\ref{eq:base}) changes to be $h^4$, which is a higher order term than the original $h^2$.
It means that we can allow larger step size $h$ by using this symmetric splitting scheme (note that $h<1$).

Similarly, we can further improve the convergence results for SVRG/SAGA-HMC by combining the symmetric splitting scheme. We give the details of the algorithms and theoretical results for SVRG2nd-HMC and SAGA2nd-HMC in the following subsections.

\subsection{SVRG2nd-HMC}

In this subsection, we propose the SVRG2nd-HMC algorithm (see Algorithm \ref{hmc-svrg-split}) by combining our SVRG-HMC (Algorithm \ref{hmc-svrg}) with the symmetric splitting scheme.

\begin{algorithm}[!h]
	\caption{SVRG2nd-HMC}\label{hmc-svrg-split}
	parameters $T, K, b, h>0$, $Dh<1$, $D\geq 1$\;
	initialize $\theta_0 = p_0 = 0$\;
	\For{$t = 0, 1, \ldots, T / K-1$}{
		compute $g = \sum_{i=1}^n \nabla f_i(\theta_{tK} + \frac{h}{2}p_{tK})$\;
		$w = \theta_{tK} + \frac{h}{2}p_{tK}$\;
		\For{$k=0, 1, \ldots, K-1$}{
			uniformly sample an index subset $I\subseteq [n]$, $|I| = b$\;
			$\tilde{\nabla}_{tK + k} = -\nabla\log\Pr(\theta_{tK + k} + \frac{h}{2}p_{tK + k}) + \frac{n}{b}\sum_{i\in I}(\nabla f_i(\theta_{tK + k} + \frac{h}{2}p_{tK + k}) - \nabla f_i(w)) + g$\;
			$p_{tK + k+1} = e^{-Dh/2} (e^{-Dh/2}p_t - h\tilde{\nabla}_{tK + k} + \sqrt{2Dh} \xi_{tK + k})$\;
			$\theta_{tK + k+1} = \theta_{tK + k} + \frac{h}{2}p_{tK + k+1} + \frac{h}{2}p_{tK + k}$\;
		}
	}
	\Return $\{\theta_t \}_{t=1}^T$\;
\end{algorithm}

The convergence result for SVRG2nd-HMC is provided in Theorem \ref{svrg2nd}. It shows that the dependency of MSE on step size $h$ can be improved from $h^2$ to $h^4$ (see (\ref{eq:svrg-hmc}) and (\ref{eq:svrg2nd-hmc})).
\begin{theorem}\label{svrg2nd}
Under Assumptions \ref{3rd} and \ref{strong}, the MSE of SVRG2nd-HMC is bounded as:
\begin{equation}	
	 \E(\hat{\phi} - \bar{\phi})^2\lesssim \min\Big\{\frac{n^2G^2}{bT}, \frac{L^2 n^2 K^2h^2}{bT}\Big(\frac{\sqrt{n^2G^2 + D^2d}}{D-L^2n^2K^2h^3b^{-1}}\Big)^2\Big\} + \frac{1}{Th} + h^4 \label{eq:svrg2nd-hmc}
\end{equation}
\end{theorem}

\subsection{SAGA2nd-HMC}

In this subsection, we propose the SAGA2nd-HMC algorithm (see Algorithm \ref{hmc-saga-split}) by combining our SAGA-HMC (Algorithm \ref{hmc-saga}) with the symmetric splitting scheme. The convergence result and algorithm details are described below.

\begin{algorithm}
	\caption{SAGA2nd-HMC}\label{hmc-saga-split}
	parameters $T, b, h > 0$, $Dh < 1$, $D\geq 1$\;
	initialize $\theta_0 = 0, p_0 = 0$\;
	initialize an array $\alpha_0^i = 0, \forall i\in [n]$\;
	compute $g = \sum_{i=1}^n \nabla f_i(\alpha_0^i)$\;
	\For{$t = 0, 1, \ldots, T-1$}{
		uniformly randomly pick a set $I\subseteq [n]$ such that $|I| = b$\;
		$\tilde{\nabla}_t = -\nabla\log \Pr(\theta_t + \frac{h}{2}p_t) + \frac{n}{b}\sum_{i\in I}(\nabla f_i(\theta_t + \frac{h}{2}p_t) - \nabla f_i(\alpha_t^i)) + g$\;
		$p_{t+1} = e^{-Dh/2} (e^{-Dh/2}p_t - h\tilde{\nabla}_t + \sqrt{2Dh}\xi_t)$\;
		$\theta_{t+1} = \theta_t + \frac{h}{2} p_{t+1} + \frac{h}{2}p_t$\;
		update $\alpha_{t+1}^i = \theta_t + \frac{h}{2}p_t$, $\forall i\in I$\;
		$g\leftarrow g + \sum_{i\in I}(\nabla f_i(\alpha_{t+1}^i) - \nabla f_i(\alpha_t^i))$\;
	}	
	\Return $\{\theta_t\}_{t=1}^T$\;
\end{algorithm}

\begin{theorem}\label{saga2nd}
	Under Assumptions \ref{3rd} and \ref{strong}, the MSE of SAGA2nd-HMC is bounded as:
\begin{equation*}
 \E(\hat{\phi} - \bar{\phi})^2\lesssim \min\Big\{\frac{n^2G^2}{bT}, \frac{L^2 n^4h^2 }{Tb^3}\Big(\frac{\sqrt{n^2G^2 + D^2d}}{D-L^2n^4h^3b^{-3}}\Big)^2\Big\} + \frac{1}{Th} + h^4
 \end{equation*}
\end{theorem}

\section{Experiment}
\label{sec:experiment}
We present experimental results in this section. We compare the proposed SVRG-HMC (Algorithm \ref{hmc-svrg}), as well as its symmetric splitting variant SVRG2nd-HMC (Algorithm \ref{hmc-svrg-split}), against SVRG-LD \citep{langevin} on Bayesian regression, Bayesian classification and Bayesian Neural Networks.
\re{The experimental results of SAGA variants (Algorithm \ref{hmc-saga} and \ref{hmc-saga-split}) are almost same as the SVRG variants. We report the corresponding SAGA experiments in Appendix \ref{app:exp}.}
In accordance with the theoretical analysis, all algorithms have fixed step size $h$, and all HMC-based algorithms have fixed friction parameter $D$; a grid search is performed to select the best step size and friction parameter for each algorithm. The minibatch size $b$ is chosen to be $10$ \re{(same as SVRG/SAGA-LD \citep{langevin})} for all algorithms, and $K$ is set to be $n/b$.

\renewcommand{\thefootnote}{\arabic{footnote}}

The experiments are tested on the real-world UCI datasets\footnote{The UCI datasets can be downloaded from \url{https://archive.ics.uci.edu/ml/datasets.html}}.
The information of the standard datasets used in our experiments are described in the following Table~\ref{tab:1} \re{and Table \ref{tab:bnn} (Section \ref{sec:bnn})}.
For each dataset (regression or classification), we partition the dataset into training (70\%), validation (10\%) and test (20\%) sets. The validation set is used to select step size as well as friction for HMC-based algorithms in an 8-fold manner.

\begin{table}[h]
\caption{Summary of standard UCI datasets for Bayesian regression and classification}
\label{tab:1}\vspace{1mm}
\centering
\begin{tabular}{cccccccc}
\hline\noalign{\smallskip}
	datasets	&	concrete	&	noise	&	parkinson	&	bike	 & pima	&	diabetic	&	eeg	\\
\noalign{\smallskip}\hline\noalign{\smallskip}
	size		&	1030		&	1503	&	5875	 &	 17379	    &	768		&	1151	&	14980			\\
	features	&	8			&	5		&	21	     &	 12    &		8	&	20			&	15		 \\
\noalign{\smallskip}\hline
\end{tabular}
\\

\vspace{1mm}
The Bayesian regression experiments were conducted on the first four UCI regression datasets, the Bayesian classification experiments were conducted on the last three UCI classification datasets\re{, and the more complicated Bayesian Neural Networks experiments were conducted on larger UCI datasets in Table \ref{tab:bnn}.}
\end{table}

\subsection{Bayesian Regression}
\label{sec:br}
In this subsection we study the performance of those aforementioned algorithms on Bayesian linear regression. Say we are provided with inputs $Z = \{(x_i, y_i)\}_{i=1}^n$ where $x_i\in \mathbb{R}^d$ and $y_i\in \mathbb{R}$. The distribution of $y_i$ given $x_i$ is modelled as $\Pr(y_i|x_i) = N(\beta\T x_i, \sigma^2)$, where the unknown parameter $\beta$ follows a prior distribution of $\N(0, I_d)$. The gradients of log-likelihood can thus be calculated as $\nabla_\beta \log\Pr(y_i| x_i, \beta) = (y_i - \beta\T x_i)x_i$ and $\nabla_\beta \log \Pr(\beta) = -\beta$.
The average test Mean-Squared Error (MSE) is reported in Figure 1.

\begin{center}
	\begin{figure}[!htb]
	\includegraphics[width=0.5\textwidth]{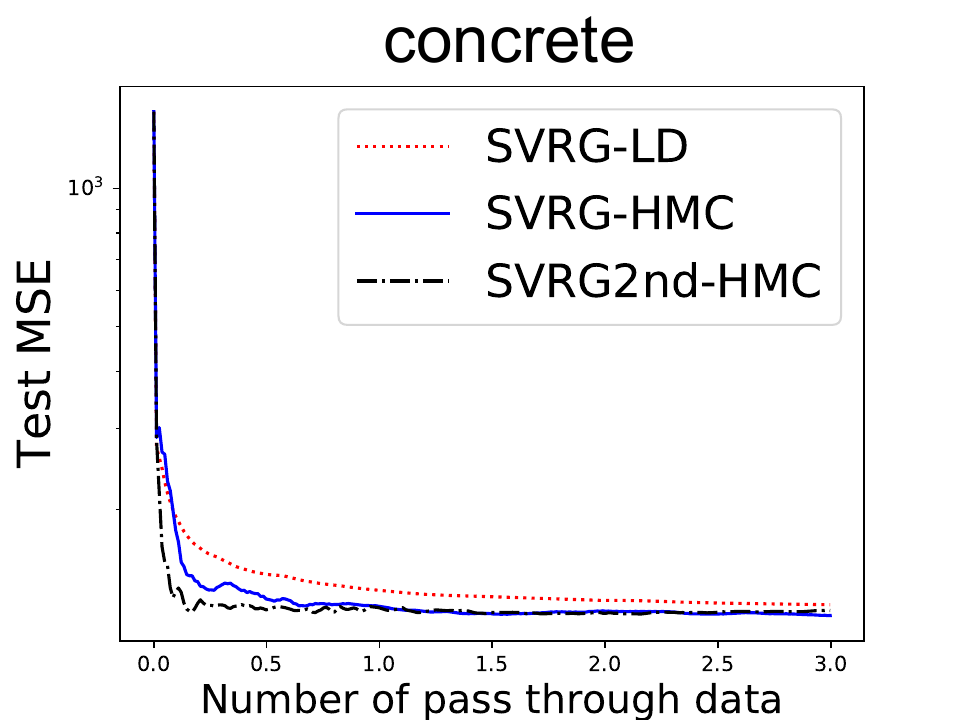}
	\includegraphics[width=0.5\textwidth]{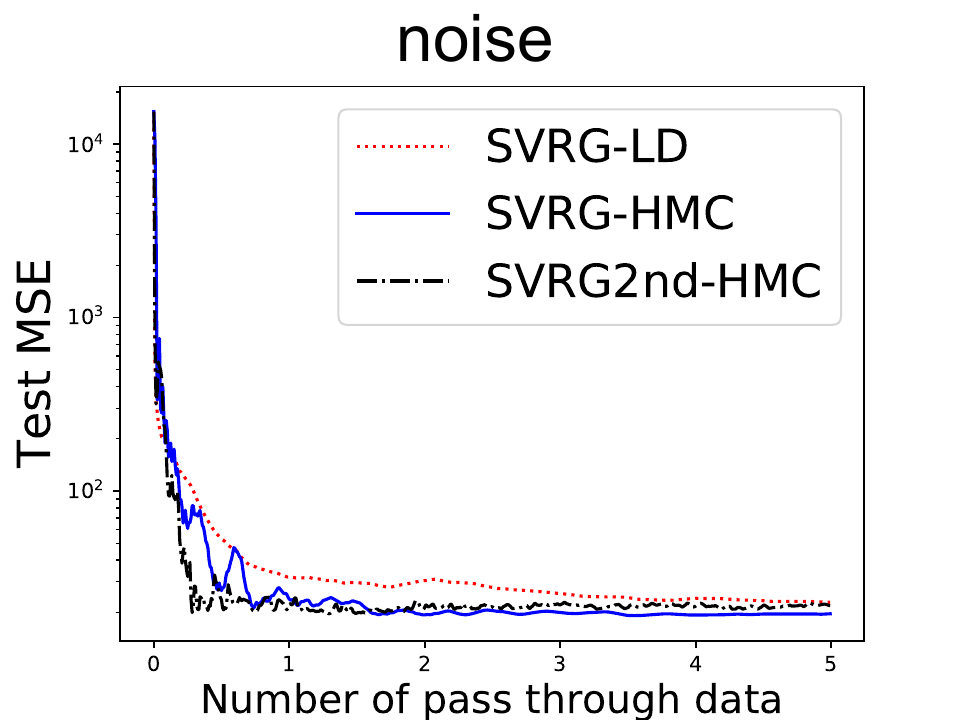}

	\includegraphics[width=0.5\textwidth]{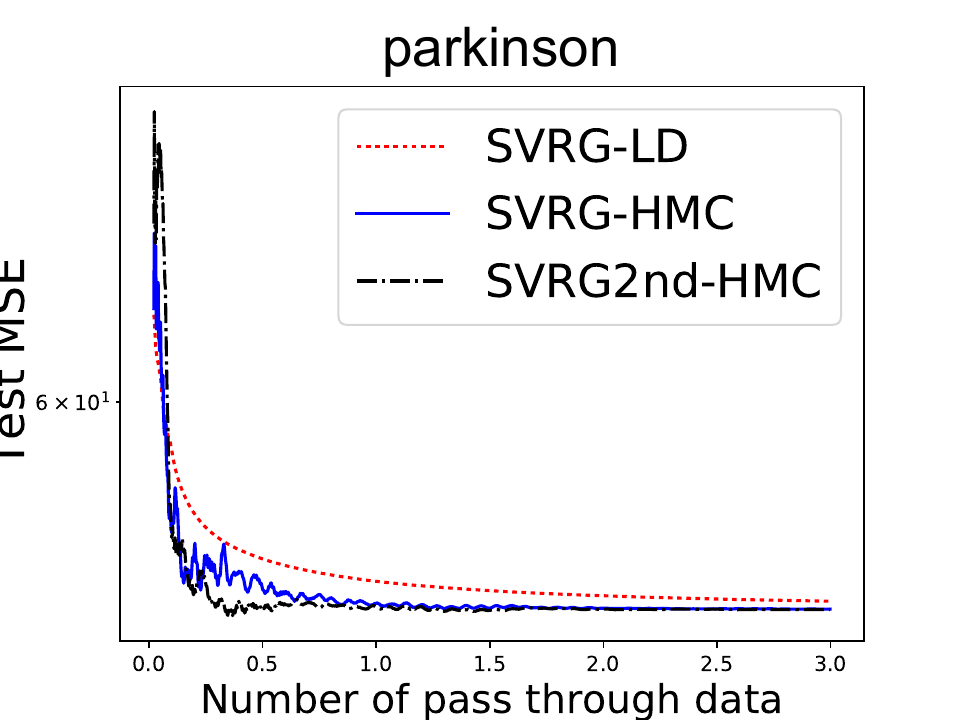}
	\includegraphics[width=0.5\textwidth]{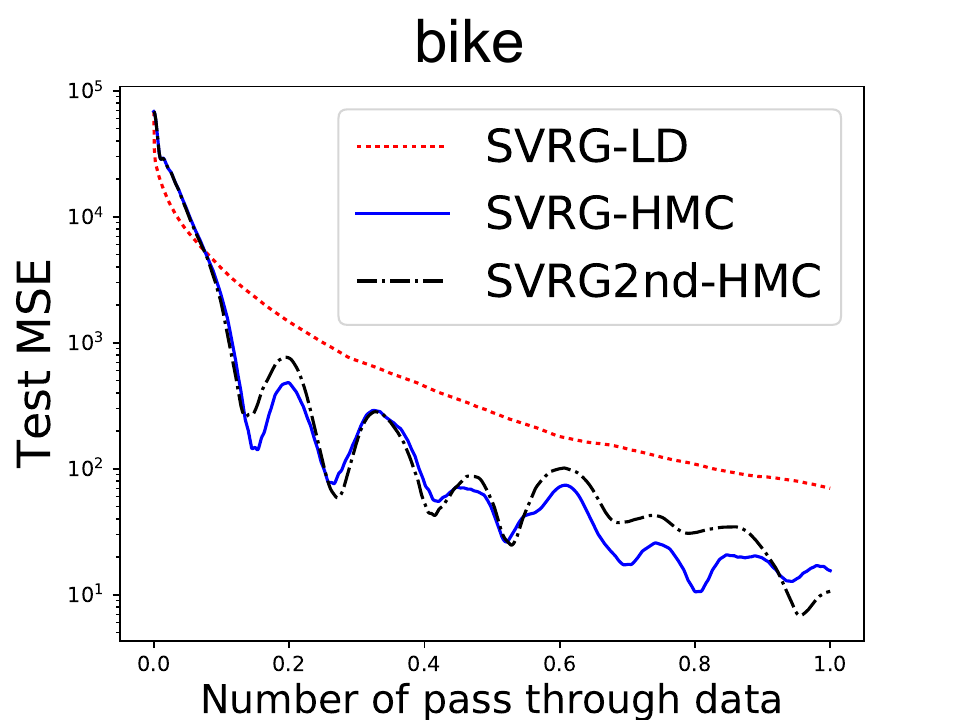}
	\caption{Performance comparison of SVRG variants on Bayesian regression tasks. The x-axis and y-axis represent number of passes through the entire training dataset and average test MSE respectively. For the bike dataset, we have omitted the first 10 MSE values from the diagram because otherwise the diagram would scale badly as MSE values are very large in the first several iterations.}   \label{fig:reg}
	\end{figure}
\end{center}

\re{As can be observed from Figure \ref{fig:reg}, SVRG-HMC as well its symmetric splitting counterpart SVRG2nd-HMC, converge markedly faster than SVRG-LD in the first pass through the whole dataset. The performance SVRG2nd-HMC is usually similar (no worse) to SVRG-HMC, and it turns out that a slightly larger step size can be chosen for SVRG2nd-HMC, which is also consistent with our theoretical results (i.e., allow larger step size).}

\subsection{Bayesian Classification}
\label{sec:bc}
\begin{center}
	\begin{figure}[!htb]
		\includegraphics[width=0.328\textwidth]{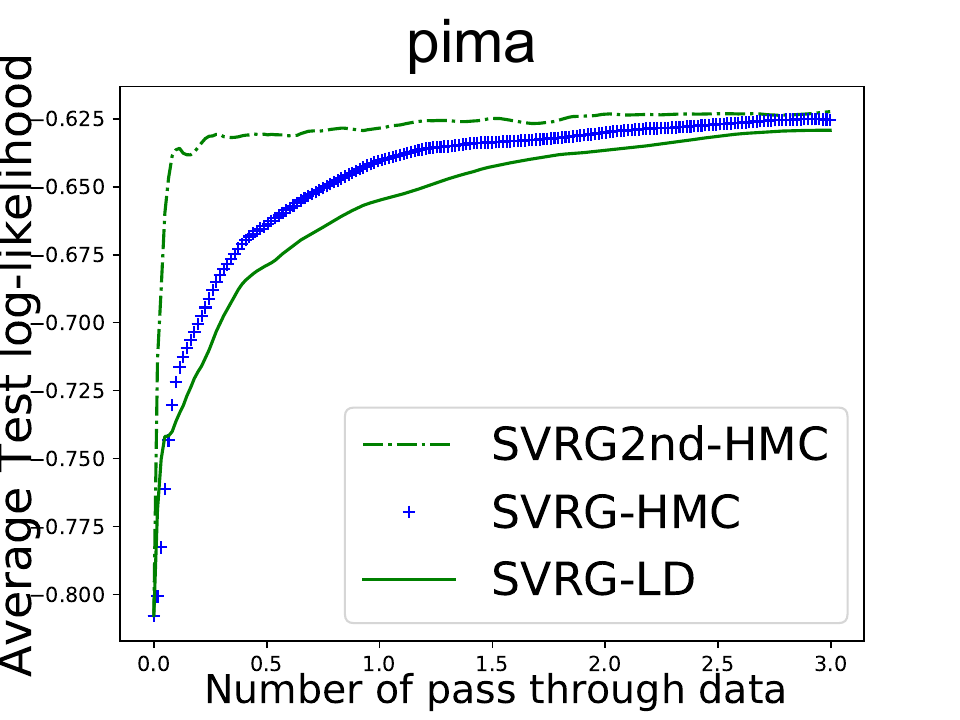}		\includegraphics[width=0.328\textwidth]{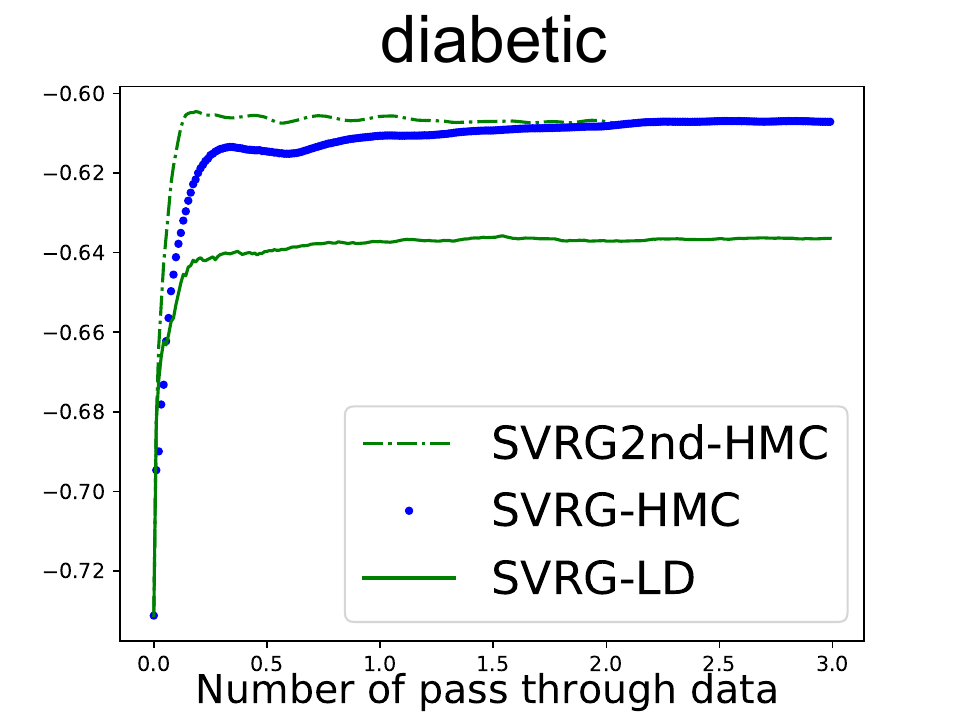}
        \includegraphics[width=0.328\textwidth]{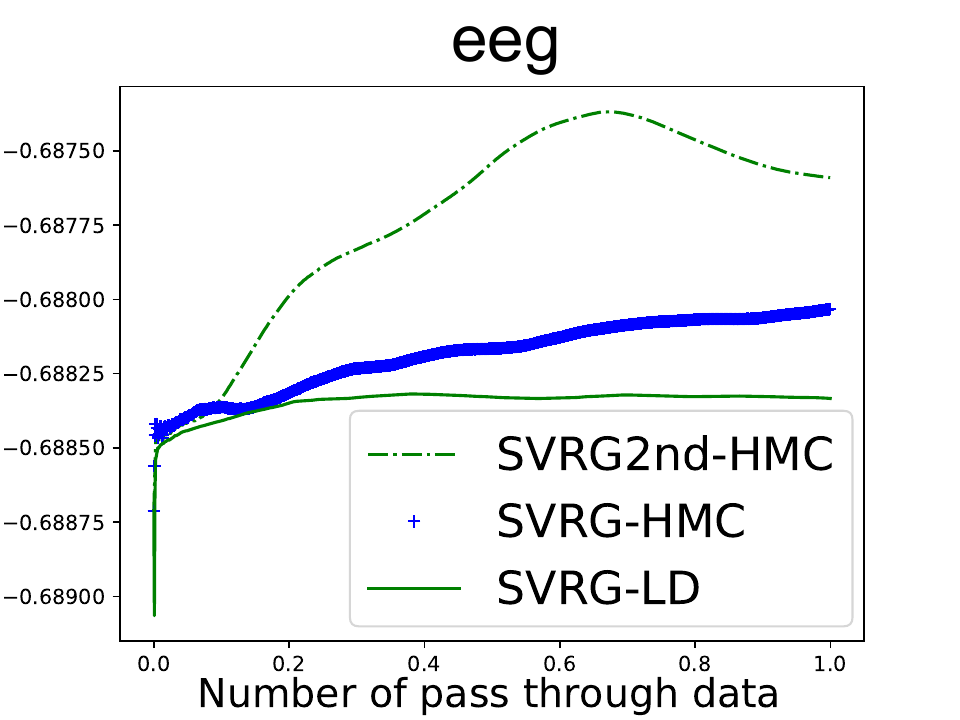}
		\caption{Performance comparison of SVRG variants on Bayesian classification tasks. The x-axis and y-axis represent number of passes through the entire training dataset and average test log-likelihood respectively.} %The experimental result on the eeg dataset is deferred to the Appendix~\ref{app:exp}.}
\label{fig:cls}
	\end{figure}
\end{center}

In this subsection we study classification tasks using Bayesian logistic classification. Suppose input data $Z = \{(x_i, y_i)\}$ where $x_i\in \mathbb{R}^d$, $y_i\in \{0, 1\}$. The distribution of the output $y_i$ is modelled as $\Pr(y_i = 1) = 1 / (1 + \exp(-\beta\T x_i))$, where the model parameter follows a prior distribution of $\N(0, I_d)$. Then the gradient of log-likelihood and log-prior can be written as $\nabla_\beta\log\Pr(y_i| x_i, \beta) = \big(y_i - 1/(1 + \exp(-\beta\T x_i))\big)x_i$ and $\nabla_\beta\log\Pr(\beta) = -\beta$.
The average test log-likelihood is reported in Figure \ref{fig:cls}.

\re{Similar to the Bayesian regression, SVRG-HMC as well its symmetric splitting counterpart SVRG2nd-HMC, converge markedly faster than SVRG-LD for the Bayesian classification tasks.
Also, the experimental results suggest that SVRG2nd-HMC converges more quickly than SVRG-HMC, which is consistent with Theorem \ref{vrhmc-svrg} and \ref{svrg2nd}.}

\re{In sum, for our four algorithms, we recommend SVRG2nd/SAGA2nd-HMC due to the better theoretical results (Theorem \ref{vrhmc-svrg}, \ref{vrhmc-saga}, \ref{svrg2nd} and \ref{saga2nd}) and practical experimental results (Figure \ref{fig:reg}--\ref{fig:sagacls}) compared with SVRG/SAGA-HMC. Further, we recommend SVRG2nd-HMC since SAGA2nd-HMC needs high memory cost and its implementation is a little bit complicated than SVRG2nd-HMC.}

\subsection{Bayesian Neural Networks}
\label{sec:bnn}
\begin{center}
	\begin{figure}[!htb]
	\includegraphics[width=0.5\textwidth]{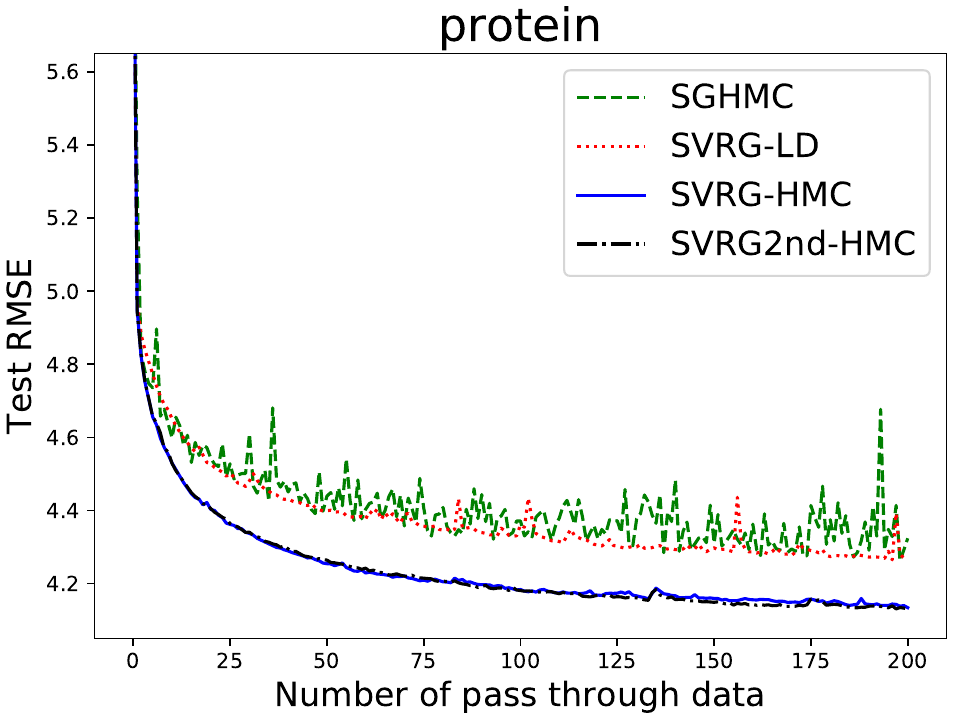}
	\includegraphics[width=0.5\textwidth]{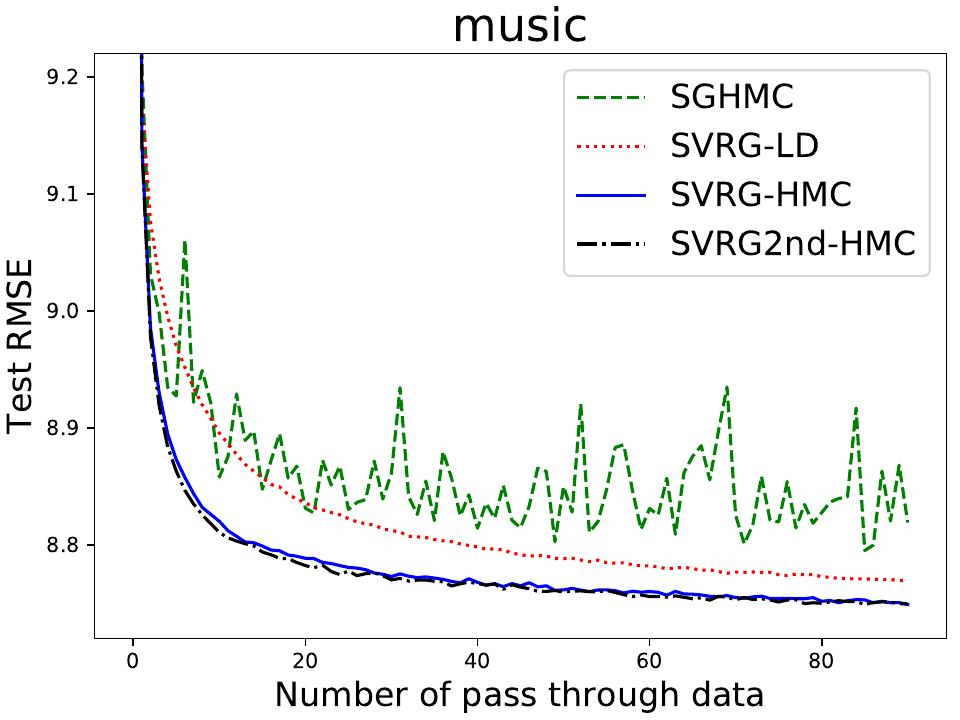}
	\caption{Performance comparison of vanilla SGHMC, SVRG-LD, SVRG-HMC, SVRG2nd-HMC on regression tasks using Bayesian neural networks. The x-axis and y-axis represent number of passes through the entire training dataset and average test RMSE respectively.}\label{fig:bnnreg}
	\end{figure}
\end{center}

\begin{center}
	\begin{figure}[!htb]
	\includegraphics[width=0.5\textwidth]{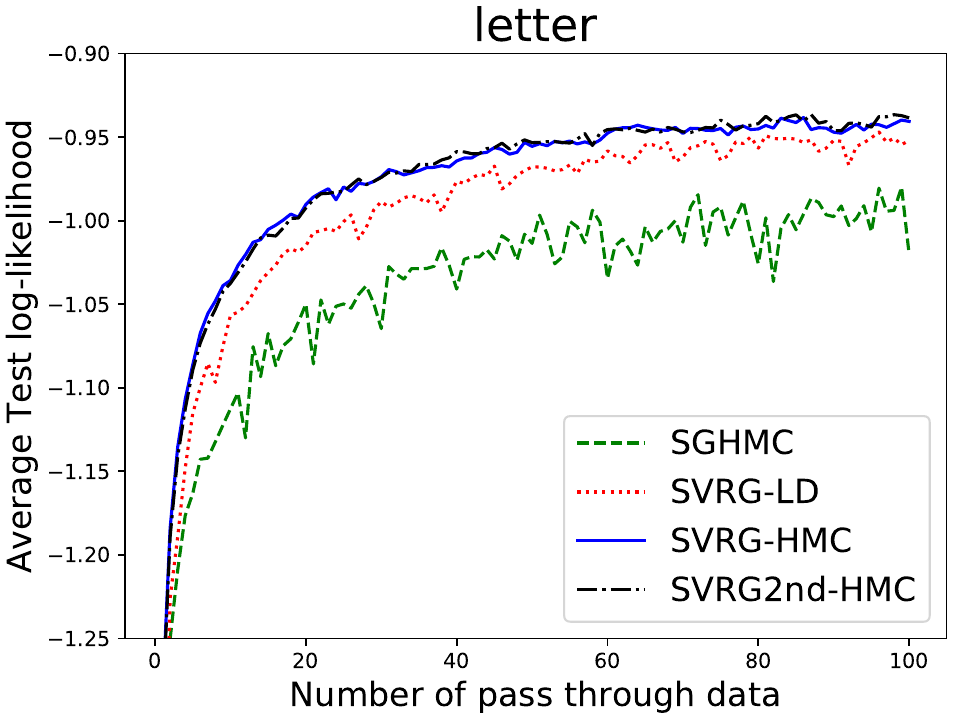}
	\includegraphics[width=0.5\textwidth]{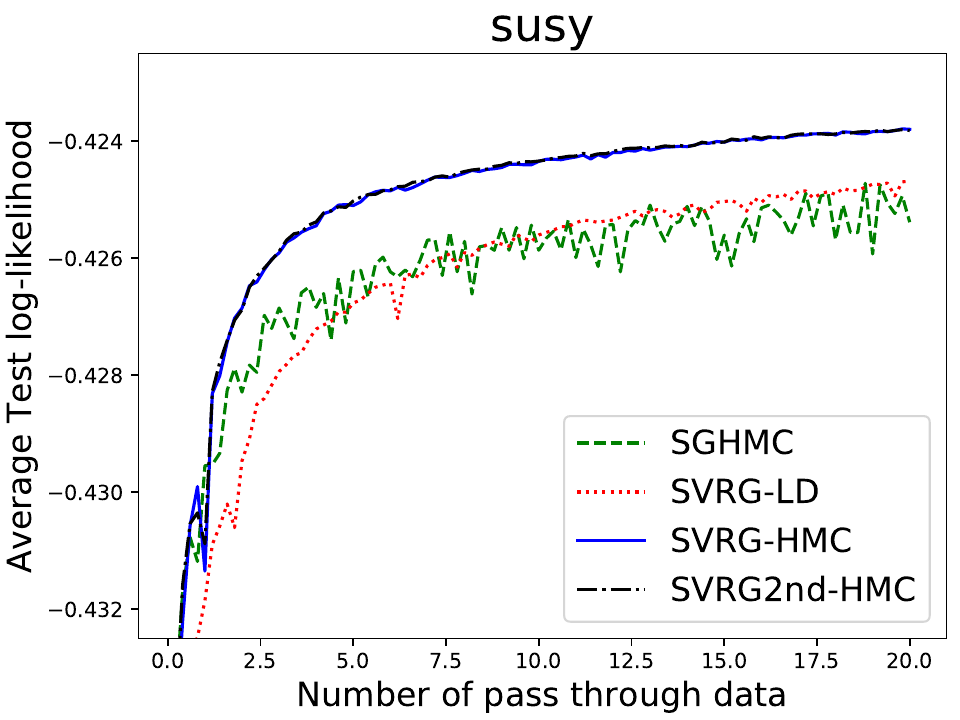}
	\caption{Performance comparison of vanilla SGHMC, SVRG-LD, SVRG-HMC, SVRG2nd-HMC on classification tasks using Bayesian neural networks. The x-axis and y-axis represent number of passes through the entire training dataset and average test log-likelihood respectively.}  \label{fig:bnncls}
	\end{figure}
\end{center}

To show the scalability of variance reduced HMC to larger datasets and its application to nonconvex problems and more complicated models, we study Bayesian neural networks tasks. In our experiments, the model is a neural network with one hidden layer which has 50 hidden units (100 hidden units for 'susy' dataset) with ReLU activation, which is denoted by $f_{NN}$. Its unknown parameter $\beta$ follows a prior distribution of $\N(0, \sigma_p^2 I_d)$. Let $t_i=f_{NN}(x_i, \beta)$ denotes output of the neural network with parameter value $\beta$ and input $x_i$. The experiments are tested on larger UCI regression and classification datasets described in Table \ref{tab:bnn}. Suppose we are provided with inputs $Z = \{(x_i, y_i)\}_{i=1}^n$ where $x_i\in \mathbb{R}^d$. In regression tasks, $y_i\in \mathbb{R}$, $t_i\in \mathbb{R}$, and the distribution of $y_i$ given $x_i$ is modelled as $\Pr(y_i|x_i) = N(t_i, \sigma_l^2)$. In binary classification tasks, $y_i\in \{0,1\}$, $t_i\in \mathbb{R}$, and $\Pr(y_i = 1) = 1 / (1 + \exp(-t_i))$. In $K$-class classification tasks ($K\geq 3$), $y_i\in \{1,2,...,K\}$, $t_i\in \mathbb{R}^K$, and $\Pr(y_i = n) = \exp(t_{in}) / \sum_{m=1}^K\exp(t_{im})$.  The code for experiments is implemented in TensorFlow. We conduct experiments for vanilla SGHMC and SVRG variants of LD and HMC algorithms. The test Root-Mean-Square Error (RMSE) for regression tasks is reported in Figure \ref{fig:bnnreg}, and the average test log-likelihood for classification tasks is reported in Figure \ref{fig:bnncls}.

\begin{table}[h]
\caption{Summary of larger UCI datasets for Bayesian neural networks experiments}
\label{tab:bnn}\vspace{1mm}
\centering
\begin{tabular}{ccccc}
\hline\noalign{\smallskip}
	datasets	&	protein	&	music	&  letter  & susy\\
\noalign{\smallskip}\hline\noalign{\smallskip}
	size		&	45730  &  515345	& 20000 & 5000000	\\			features	&	9			&  90	& 16 & 18 \\
\noalign{\smallskip}\hline
\end{tabular}
\\

\vspace{1mm}
The Bayesian neural network regression experiments were conducted on the first two UCI regression datasets and the classification experiments were conducted on the last two UCI classification datasets. The 'letter' dataset is 26-class and the 'susy' dataset is binary class.
\end{table}

\vspace{-3mm}

\vspace{-3mm}

Experimental results show that SVRG/SVRG2nd-HMC outperforms vanilla SGHMC and SVRG-LD, often by a significant gap.
In particularly, this means that variance reduction technique indeed helps the convergence of SGHMC, i.e., the performance gap between SVRG/SVRG2nd-HMC and SVRG-LD in Figure \ref{fig:reg}--\ref{fig:bnncls} is not only coming from the superiority of HMC compared with LD.
Similar to previous Section \ref{sec:br} and \ref{sec:bc}, the performance SVRG2nd-HMC is usually similar (no worse) to SVRG-HMC, and our experiments found sometimes a slightly larger step size can be chosen for SVRG2nd-HMC (while the same step size brings SVRG-HMC to NaN), which is also consistent with our theoretical results Theorem \ref{svrg2nd}.

\section{Conclusion}
In this paper, we propose four variance-reduced Hamiltonian Monte Carlo algorithms, i.e., SVRG-HMC, SAGA-HMC, SVRG2nd-HMC and SAGA2nd-HMC for Bayesian Inference.
These proposed algorithms guarantee improved theoretical convergence results and converge markedly faster than the benchmarks (vanilla SGHMC and SVRG/SAGA-LD) in practice. \re{In conclusion, the SVRG2nd/SAGA2nd-HMC are more preferable than SVRG/SAGA-HMC according to our theoretical and experimental results.}
We would like to note that, our variance-reduced Hamiltonian Monte Carlo samplers are not Markovian procedures, but fortunately our theoretical analysis does not rely on any properties of Markov processes, and so it does not affect the correctness of Theorem \ref{vrhmc-svrg}, \ref{vrhmc-saga}, \ref{svrg2nd} and \ref{saga2nd}.

For future work, \re{it would be interesting to study whether our analysis can be apply to vanilla SGHMC without variance reduction. To the best of our knowledge, there is no existing work to prove that SGHMC is better than SGLD. On the other hand,} we note that stochastic thermostat \citep{ding2014bayesian} could outperform both SGLD and SGHMC. It might be interesting to study if a variance-reduced variant of stochastic thermostat could also beat SVRG-LD and SVRG/SVRG2nd-HMC both theoretically and experimentally.

\section*{Acknowledgments}
We would like to thank Chang Liu for useful discussions.

\bibliographystyle{plainnat}
\bibliography{ref}

\newpage
\appendix

\section{SAGA Experiments}
\label{app:exp}
In this appendix, we report the corresponding experimental results of SAGA variants (i.e., SAGA-LD, SAGA-HMC and SAGA2nd-HMC) for Bayesian regression and Bayesian classification tasks. The settings are the same as those in Section~\ref{sec:br} and \ref{sec:bc}.
\begin{center}\vspace{-3mm}
	\begin{figure}[!htb]
	\includegraphics[width=0.5\textwidth]{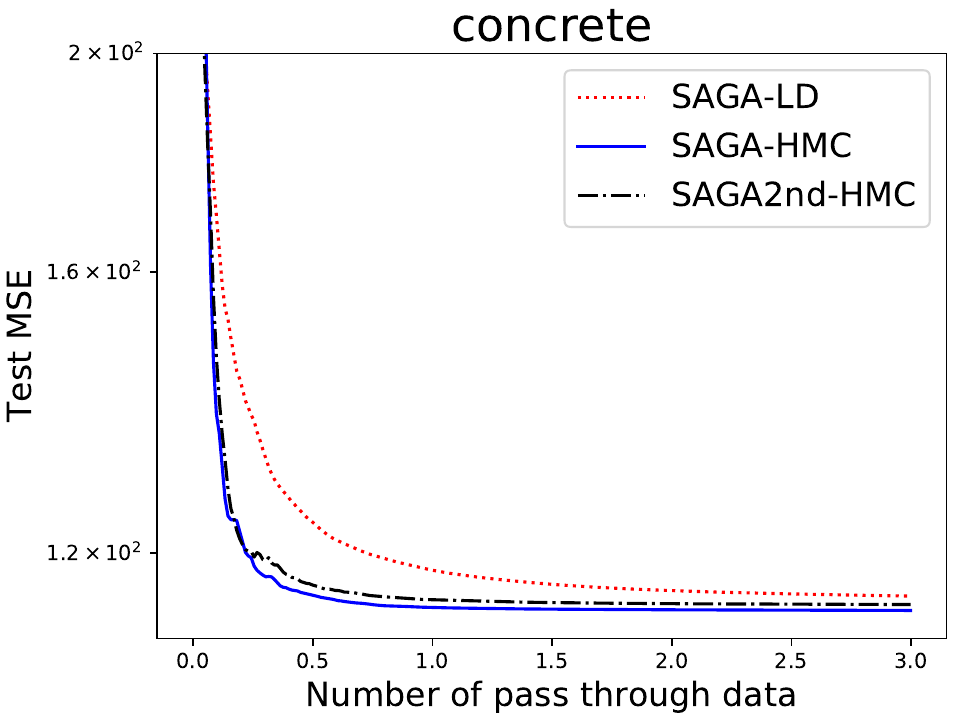}
	\includegraphics[width=0.5\textwidth]{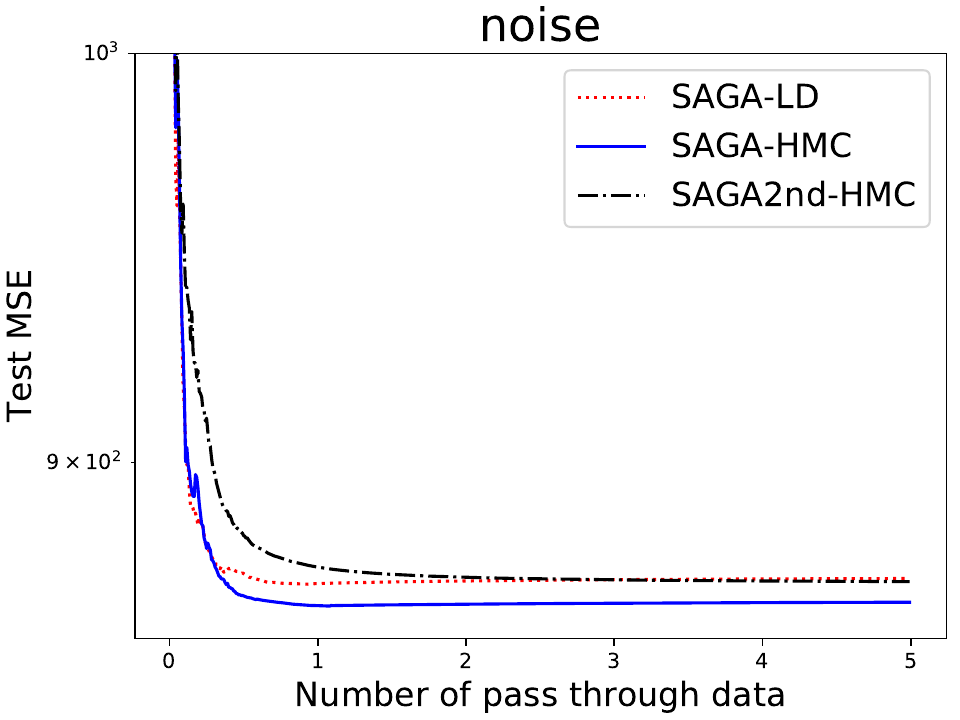}

	\includegraphics[width=0.5\textwidth]{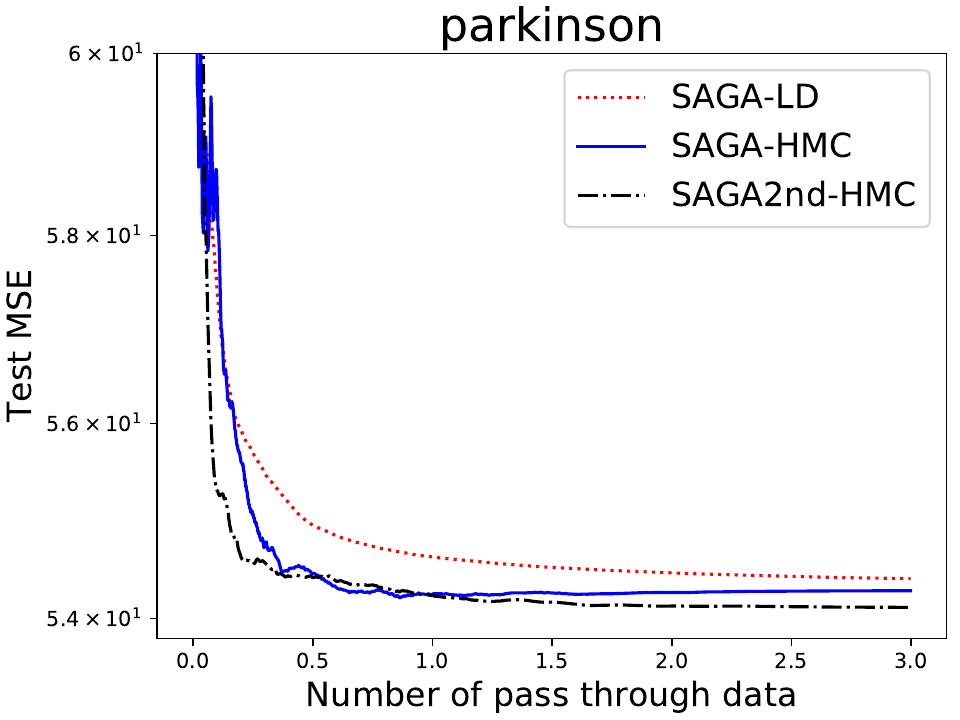}
	\includegraphics[width=0.5\textwidth]{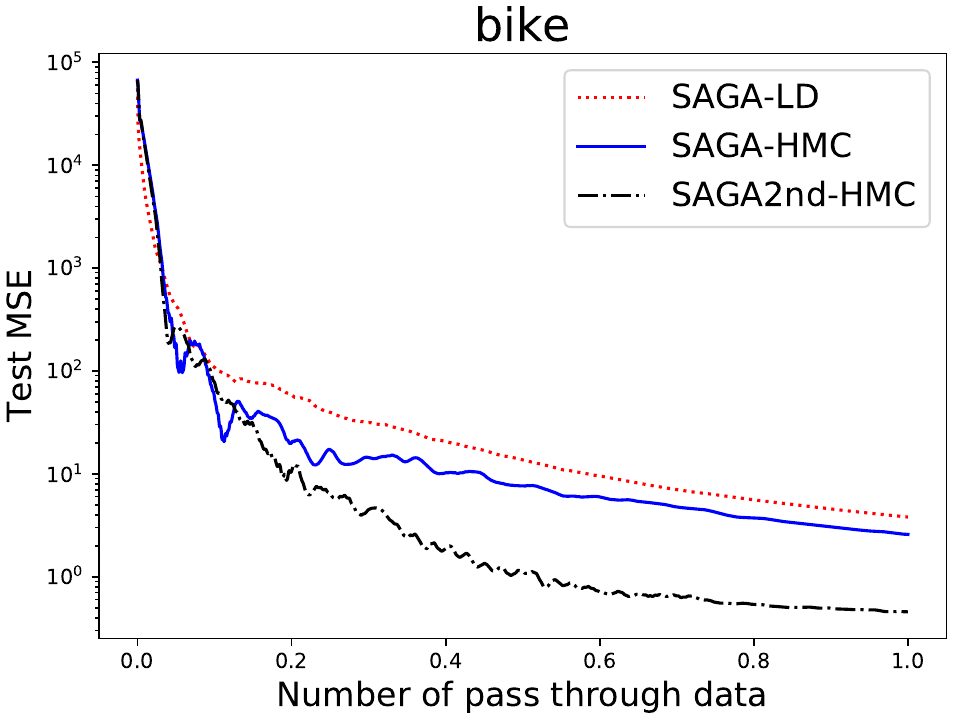}\vspace{-2mm}
	\caption{Performance comparison of SAGA variants on Bayesian regression tasks.}    \label{fig:sagareg}
	\end{figure}
\end{center}\vspace{-4mm}
\begin{center}
	\begin{figure}[!htb]
		\includegraphics[width=0.328\textwidth]{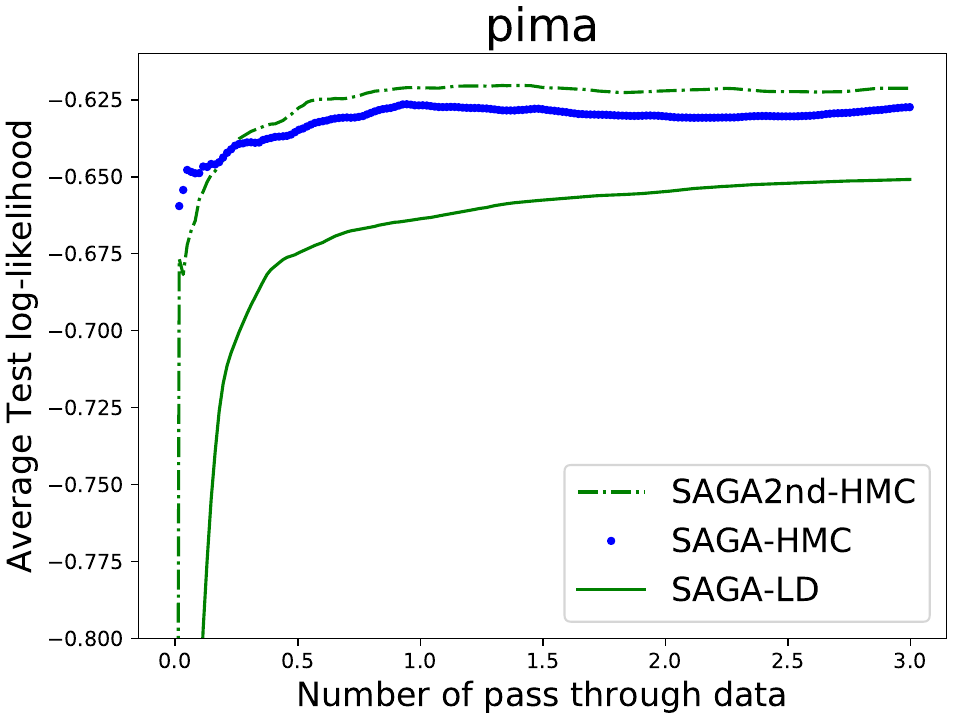}		\includegraphics[width=0.328\textwidth]{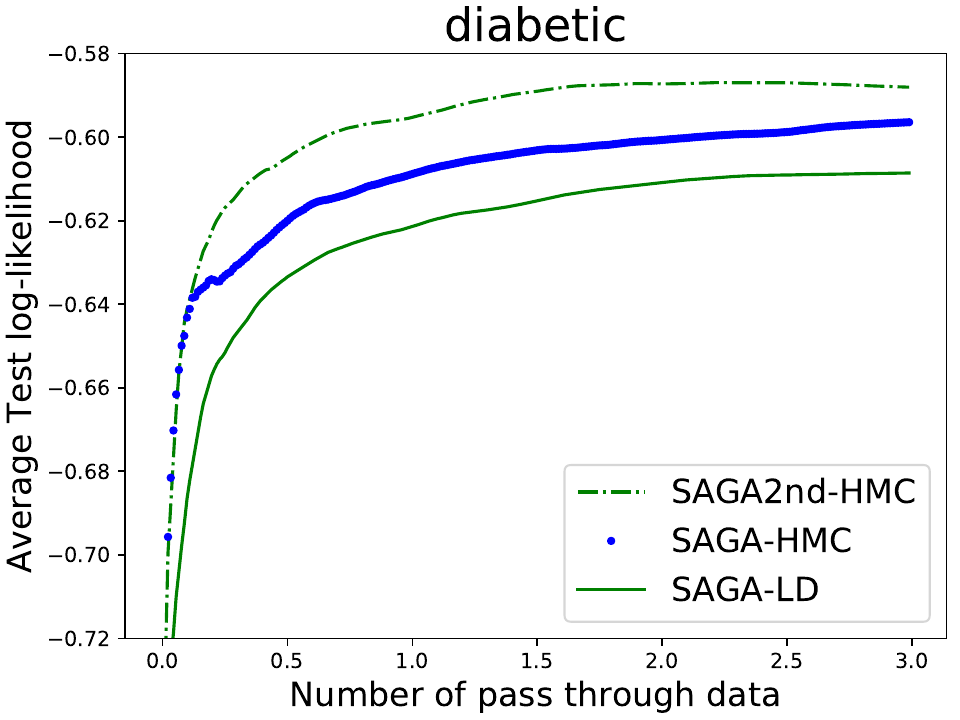}
        \includegraphics[width=0.328\textwidth]{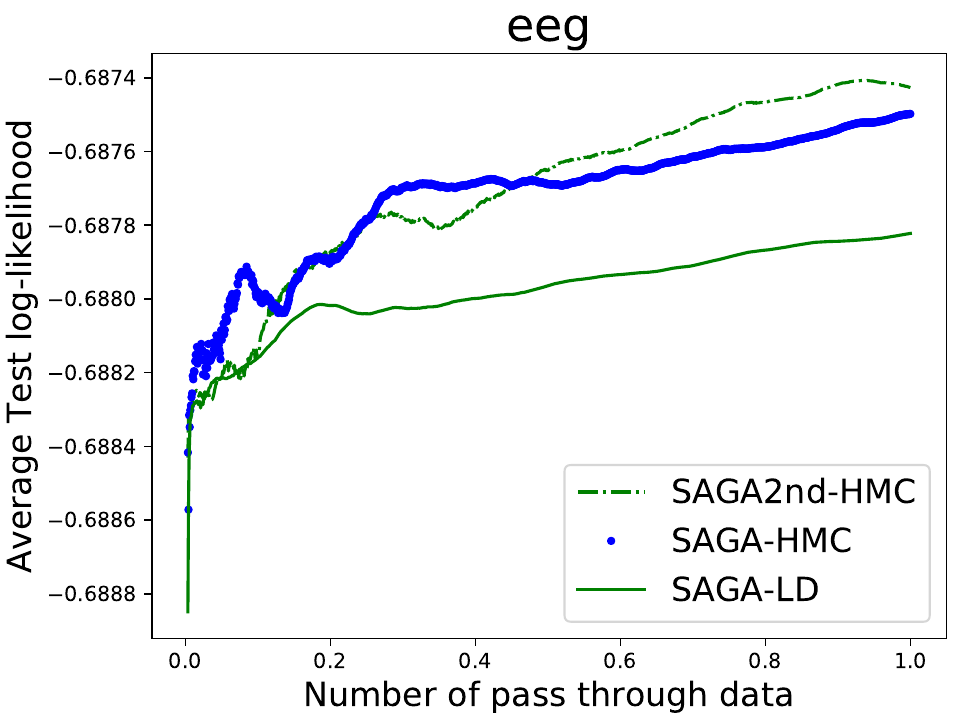}\vspace{-2mm}
		\caption{Performance comparison of SAGA variants on Bayesian classification tasks.}
    \label{fig:sagacls}
	\end{figure}
\end{center}

\section{Missing Proofs}
\label{app:proofs}
In this appendix, we provide the detailed proofs for Corollary \ref{sghmc}, Theorem \ref{vrhmc-svrg}, Lemma \ref{lm:comp}, and Theorem \ref{vrhmc-saga}, \ref{svrg2nd} and \ref{saga2nd}.

\subsection{Proof of Corollary \ref{vrhmc-svrg} }
To prove this corollary, it is sufficient to show $\E\|\tilde{\nabla}_t - \nabla f(\theta_t)\|^2\leq n^2G^2/b$.
Recall that $\tilde{\nabla}_t = \frac{n}{b}\sum_{i\in I} \nabla f_i(\theta_t)$, $I$ being a $b$-element index set uniformly randomly drawn (with replacement) from $\{1, 2, \ldots, n\}$
and $\nabla f(\theta_t)=\sum_{j=1}^n \nabla f_j(\theta_t)$.
Now, we prove this inequality as follows:
\begin{align*}
\E_I\|\tilde{\nabla}_t - \nabla f(\theta_t)\|^2
&=\E_I\big\|\frac{n}{b}\sum_{i\in I} \nabla f_i(\theta_t) -\sum_{j=1}^n \nabla f_j(\theta_t) \big\|^2 \notag \\
&=n^2\E_I\big\|\frac{1}{b}\sum_{i\in I} \nabla f_i(\theta_t) -\frac{1}{n}\sum_{j=1}^n \nabla f_j(\theta_t) \big\|^2 \notag \\
&=n^2\E_I\big\|\frac{1}{b}\sum_{i\in I} \Big(\nabla f_i(\theta_t) -\frac{1}{n}\sum_{j=1}^n \nabla f_j(\theta_t)\Big) \big\|^2 \notag \\
&=\frac{n^2}{b^2}\E_I\big\|\sum_{i\in I} \Big(\nabla f_i(\theta_t) -\frac{1}{n}\sum_{j=1}^n \nabla f_j(\theta_t)\Big) \big\|^2 \notag \\
&=\frac{n^2}{b^2}\E_I\big\|\sum_{i\in I} \Big(\nabla f_i(\theta_t) -\frac{1}{n}\sum_{j=1}^n \nabla f_j(\theta_t)\Big) \big\|^2 \notag \\
&=\frac{n^2}{b}\E_i\big\|\nabla f_i(\theta_t) -\frac{1}{n}\sum_{j=1}^n \nabla f_j(\theta_t) \big\|^2 \notag \\
&\leq\frac{n^2}{b}\E_i\big\|\nabla f_i(\theta_t)\big\|^2  \\
&\leq \frac{n^2 G^2}{b}
\end{align*}
where the last two inequalities hold since $\E(X - \E X)^2\leq \E X^2$ for any random variable $X$ and $f_i$ is $G$-Lipschitz.
\qed

\subsection{Proof of Theorem \ref{vrhmc-svrg}}

According to Corollary \ref{noise}, we have:
\begin{equation}\label{eq:corr}
\mathbb{E}[(\hat{\phi} - \bar{\phi})^2]\lesssim \frac{1}{T^2}\sum_{t=0}^{T-1} \mathbb{E}[\|\Delta_t\|^2] + \frac{1}{Th} + h^2
\end{equation}
where $\Delta_t = \tilde{\nabla}_{t} - \nabla f(\theta_t)$ is the additive error in estimating the full gradient $\nabla f(\theta_t)$. By applying the variance reduction technique, we need to upper bound the summation $\sum_{t=1}^T\mathbb{E}[\|\Delta_t\|^2]$.

Unpacking the definition of $\Delta_t$ and $\tilde{\nabla}_t$, we have:
\begin{align}
& \sum_{t=0}^{T-1} \E[\|\Delta_t\|^2] \notag \\
=& \sum_{t=0}^{T-1} \E[\|-\nabla\log\Pr(\theta_t) + \frac{n}{b}\sum_{i\in I}(\nabla f_i(\theta_t) - \nabla f_i(\theta_{\lfloor\frac{t}{K}\rfloor K})) + g - \nabla f(\theta_t) \|^2] \notag \\
=& \sum_{t=0}^{T-1} n^2\E[\|\frac{1}{b}\sum_{i\in I}(\nabla f_i(\theta_t) - \nabla f_i(\theta_{\lfloor\frac{t}{K}\rfloor K})) - \frac{1}{n}\sum_{j=1}^n (\nabla f_j(\theta_t) - \nabla f_j(\theta_{\lfloor\frac{t}{K}\rfloor K}))\|^2] \notag\\
\leq& \frac{n^2}{b}\sum_{t=0}^{T-1}\E_{i\in[n]}\|\nabla f_i(\theta_t) - \nabla f_i(\theta_{\lfloor\frac{t}{K}\rfloor K})\|^2 \label{eq:first}
\end{align}
The Inequality (\ref{eq:first}) is due to $\E(X - \E X)^2\leq \E X^2$ for any random variable $X$. In the rightmost summation, index $i$ is picked uniformly random from $[n] = \{1, 2, \ldots, n\}$.

Then, we bound the RHS of (\ref{eq:first}) as follows:
\begin{align}
\frac{n^2}{b}\sum_{t=0}^{T-1}\E_{i\in [n]}\|\nabla f_i(\theta_t) - \nabla f_i(\theta_{\lfloor\frac{t}{K}\rfloor K})\|^2
&\leq \frac{n^2}{b}\sum_{t=0}^{T-1}L^2\mathbb{E}\|\theta_t - \theta_{\lfloor\frac{t}{K}\rfloor K}\|^2 \notag\\
&\leq \frac{L^2 n^2}{b}\sum_{t=0}^{T-1} K\sum_{j=\lfloor\frac{t}{K}\rfloor K }^{t-1}\E\|\theta_{j+1} - \theta_j\|^2 \notag\\
&\leq \frac{L^2 n^2 K^2}{b}\sum_{t=0}^{T-1}\mathbb{E}\|\theta_{t+1} - \theta_t\|^2 \label{eq:tmp}
\end{align}

The first inequality is by $L$-smoothness of all $f_i$'s, and the second one is by Cauchy's inequality.

By our algorithm, $\|\theta_{t+1} - \theta_t\|^2 = h^2\|p_{t+1}\|^2$, so we need to upper bound $\mathbb{E}\|p_{t+1}\|^2$ for each $0\leq t<T$.

By the recursion of $p_{t+1}$,
$$\begin{aligned}
& \mathbb{E}\|p_{t+1}\|^2 \\
=& \mathbb{E}\|(1-Dh)p_t - h\tilde{\nabla}_t + \sqrt{2Dh}\xi_t\|^2\\
=& \mathbb{E}\|(1-Dh)p_t - h\nabla f(\theta_t) - h\Delta_t + \sqrt{2Dh}\xi_t\|^2\\
=& \mathbb{E}\|(1-Dh)p_t - h\nabla f(\theta_t)\|^2 + h^2\mathbb{E}[\|\Delta_t\|^2] + 2Dhd\\
\leq& (1-Dh)^2\mathbb{E}\|p_t\|^2 + 2Gnh(1-Dh)\sqrt{\mathbb{E}\|p_t\|^2} + h^2n^2G^2 \\
&\quad + h^2\mathbb{E}[\|\Delta_t\|^2] + 2Dhd
\end{aligned}$$

The third equality holds because $\E[\Delta_t] = \E[\xi_t] = 0$ and $\Delta_t$ and $\xi_t$ are independent. The first inequality takes advantage of $\|\nabla f\|\leq nG$ and $\E\|p_t\|\leq \sqrt{\E\|p_t\|^2}$.

Define $S = \sum_{t=1}^T\mathbb{E}[\|p_t\|^2]$. Then, taking a grand summation over $t = 0, 1, \ldots, T-1$,
$$\begin{aligned}
S&\leq (1-Dh)^2 S + 2nGh(1-Dh)\sum_{t=0}^{T-1}\sqrt{\mathbb{E}\|p_t\|^2}\\
&\qquad + T(h^2n^2G^2 + 2Dhd) +  h^2\sum_{t=0}^{T-1}\mathbb{E}[\|\Delta_t\|^2]\\
&\leq (1-Dh)^2 S + 2nGh(1-Dh)\sqrt{T} \sqrt{S}\\
&\qquad + T(h^2n^2G^2 + 2Dhd) + \frac{L^2 n^2 K^2h^4}{b}S
\end{aligned}$$
The second inequality again contains an implicit Cauchy's inequality.

Rearranging the terms we have:
$$\begin{aligned}
(1 - (1-Dh)^2 - \frac{L^2n^2K^2h^4}{b})\frac{S}{T} - 2nGh(1-Dh)\sqrt{\frac{S}{T}} - (h^2n^2G^2 + 2Dhd)\leq 0
\end{aligned}$$

Solving a quadratic equation with respect to $\sqrt{S / T}$ and ignoring constant factors, we have:
\begin{equation}\label{eq:tmp2}
\sqrt{\frac{S}{T}} \lesssim \frac{nG + \sqrt{n^2G^2 + D^2d}}{D-L^2n^2K^2h^3b^{-1}}
\lesssim \frac{\sqrt{n^2G^2 + D^2d}}{D-L^2n^2K^2h^3b^{-1}}
\end{equation}

From (\ref{eq:first}) and (\ref{eq:tmp}), we have:
$$\sum_{t=0}^{T-1} \mathbb{E}[\|\Delta_t\|^2]\leq \frac{L^2 n^2 K^2}{b}\sum_{t=0}^{T-1}\mathbb{E}\|\theta_{t+1} - \theta_t\|^2$$

Recall that $\|\theta_{t+1} - \theta_t\|^2 = h^2\|p_{t+1}\|^2$ and $S=\sum_{t=1}^T\mathbb{E}[\|p_t\|^2]$, we have:
\begin{equation}\label{eq:part1}
\frac{1}{T}\sum_{t=1}^T \mathbb{E}[\|\Delta_t\|^2] \lesssim \frac{L^2 n^2 K^2h^2}{b}\left(\frac{\sqrt{n^2G^2 + D^2d}}{D-L^2n^2K^2h^3b^{-1}}\right)^2
\end{equation}

On the other hand, we can bound (\ref{eq:first}) as follows:
\begin{align}
\sum_{t=0}^{T-1} \E[\|\Delta_t\|^2]
&\leq \frac{n^2}{b}\sum_{t=0}^{T-1}\E_{i\in[n]}\|\nabla f_i(\theta_t) - \nabla f_i(\theta_{\lfloor\frac{t}{K}\rfloor K})\|^2 \notag \\
&\leq \frac{4Tn^2G^2}{b} \label{eq:part2}
\end{align}
where (\ref{eq:part2}) holds due to all $f_i$'s are $G$-Lipschitz and Cauchy's inequality $\|a+b\|^2\le 2(\|a\|^2+\|b\|^2)$.

Now, the proof of Theorem \ref{vrhmc-svrg} is finished by combining (\ref{eq:corr}), (\ref{eq:part1}) and (\ref{eq:part2}).
\qed

\vspace{2mm}
\subsection{Proof of Lemma \ref{lm:comp}}

Since $G^2 > K^2(n^2L^2 h^2G^2 + hd) > K^2n^2L^2h^2G^2$, we have $h < \frac{1}{nKL}$.
Therefore, we have
\begin{align*}
D - \frac{h^3L^2n^2K^2}{b}
&> D - \frac{1}{n^3K^3L^3}\frac{L^2n^2K^2}{b} \notag\\
&= D - \frac{1}{nKbL} \notag\\
&\gg D - 0.1 \geq 0.9 D \notag
\end{align*}
where the last line holds since $nKbL\gg 10$ and $D>1$ (see Line 1 of Algorithm \ref{hmc-svrg}).
Thus, the proof is reduced to comparing $h^2L^2n^2G^2 + hd$ and $\frac{h^2L^2 n^2 G^2}{D^2} + h^2L^2d$ asymptotically.

Clearly,
$$\begin{aligned}
&\frac{h^2L^2 n^2 G^2}{D^2} + h^2L^2d \\
&\leq \frac{h^2L^2 n^2 G^2}{D^2} + \frac{1}{nKL}hL^2 d \\
&= \frac{h^2L^2 n^2 G^2}{D^2} + hd\frac{L}{nK}\\
&\leq \max\{\frac{1}{D^2}, \frac{L}{nK}\}(h^2L^2n^2G^2 + hd)
\end{aligned}$$
Note that if $D\ge {1}/{L\sqrt{h}}$, then $\max\{\frac{1}{D^2}, \frac{L}{nK}\}$ turns to be $\frac{L}{nK}$ which is very small. The reason is that:
\begin{align*}
D^2&\ge\frac{1}{L^2h}
\ge\frac{1}{L^2\frac{1}{nKL}} = \frac{nK}{L}
\end{align*}
where the second inequality holds due to $h < \frac{1}{nKL}$ which is mentioned above.
\qed

\vspace{3mm}
\subsection{Proof of Theorem \ref{vrhmc-saga}}
Defining $\Delta_t = \tilde{\nabla}_t - \nabla f(\theta_t)$, it suffices to upper bound $\sum_{t=0}^{T-1}\E[\|\Delta_t\|^2]$ according to (\ref{eq:corr}). Unpacking the definition of $\Delta_t$ and $\tilde{\nabla}_t$, we have:
\begin{align}
\sum_{t=0}^{T-1}\E[\|\Delta_t\|^2]
&= \sum_{t=0}^{T-1}\E[\|\tilde{\nabla}_t - \nabla f(\theta_t)\|^2]\notag\\
&= \sum_{t=0}^{T-1}\E[\|\frac{n}{b}\sum_{i\in I}(\nabla f_i(\theta_t) - \nabla f_i(\alpha_t^i)) -\sum_{j=1}^n(\nabla f_j(\theta_t) - \nabla f_j(\alpha_t^j))\|^2]\notag\\
&= \sum_{t=0}^{T-1}n^2 \E[\|\frac{1}{b}\sum_{i\in I}(\nabla f_i(\theta_t) - \nabla f_i(\alpha_t^i)) - \frac{1}{n}\sum_{j=1}^n(\nabla f_j(\theta_t) - \nabla f_j(\alpha_t^j))  \|^2]\notag\\
&\leq \frac{n^2}{b}\sum_{t=0}^{T-1}\E_{i\in [n]}[\|\nabla f_i(\theta_t) - \nabla f_i(\alpha_t^i)\|^2] \label{eq:sagafirst}\\
&\leq \frac{L^2 n^2}{b}\sum_{t=0}^{T-1}\E_{i\in [n]}[\|\theta_t - \alpha_t^i\|^2]\notag
\end{align}
The first inequality is because $\E(X - \E X)^2\leq \E X^2$ for any random variable $X$; the second inequality holds due to $L$-smoothness of $f_i$'s.

Let $\gamma = 1 - (1 - 1/n)^b$. Next we upper bound each $\E[\|\theta_t - \alpha_t^i\|^2]$ in the following manner.
$$\begin{aligned}
\E[\|\theta_t - \alpha_t^i\|^2] &= \sum_{j=0}^{t-1}\E[\|\theta_t - \theta_j\|^2] \Pr(\alpha_t^i = \theta_j)\\
&= \sum_{j=0}^{t-1}\E[\|\theta_t - \theta_j\|^2](1-\gamma)^{t-j-1}\gamma\\
&= h^2\sum_{j=0}^{t-1} \E[\|p_t + p_{t-1} + \cdots + p_{j+1}\|^2] (1-\gamma)^{t-j-1}\gamma\\
&\leq h^2\sum_{j=0}^{t-1} (t-j) (1-\gamma)^{t-j-1}\gamma(\E[\|p_t\|^2] + \E[\|p_{t-1}\|^2] + \cdots + \E[\|p_{j+1}\|^2])\\
&= h^2\sum_{j=1}^{t}\E[\|p_j\|^2] \sum_{k=0}^{j-1}(t-k)(1-\gamma)^{t-k-1}\gamma\\
&\leq h^2\sum_{j=1}^{t}\E[\|p_j\|^2] \sum_{k=0}^\infty (t-j+k+1)(1-\gamma)^{t-j+k}\gamma\\
&< h^2\sum_{j=1}^t \E[\|p_j\|^2] (\frac{1}{\gamma} + t - j) (1-\gamma)^{t-j}
\end{aligned}$$
The second equality is by direct calculation $\Pr(\alpha_t^i = \theta_j) = (1-\gamma)^{t-j-1}\gamma$; the first inequality is a direct application of Cauchy's inequality; the last inequality is a weighted summation of geometric series $(1-\gamma)^{t-j+k}, k\geq 0$.

Summing over all $t$ and $i$, we then have:
$$\begin{aligned}
\sum_{t=0}^{T-1}\E_{i\in[n]}[\|\theta_t - \alpha_t^i\|^2] &\leq h^2\sum_{t=0}^{T-1}\sum_{j=1}^t \E[\|p_j\|^2] (\frac{1}{\gamma} + t - j) (1-\gamma)^{t-j}\\
&\leq h^2\sum_{t=1}^{T-1} \E[\|p_t\|^2]\sum_{j=t}^{T-1} (\frac{1}{\gamma} + j-t)(1-\gamma)^{j-t}\\
&\leq h^2\sum_{t=1}^{T-1} \E[\|p_t\|^2]\sum_{j=t}^\infty (\frac{1}{\gamma} + j-t)(1-\gamma)^{j-t}\\
&\leq h^2\sum_{t=1}^{T-1} \E[\|p_t\|^2]\frac{2}{\gamma^2} \\
&= \frac{2}{\gamma^2}h^2\sum_{t=1}^{T-1} \E[\|p_t\|^2]\\
&\leq \frac{8h^2n^2}{b^2}\sum_{t=1}^{T} \E[\|p_t\|^2]
\end{aligned}$$
The last inequality is because $(1-1/n)^b \leq \frac{1}{1 + \frac{b}{n-1}}$, and thus $\gamma = 1 - (1-1/n)^b \geq \frac{\frac{b}{n-1}}{1 + \frac{b}{n-1}} > \frac{b}{2n}$; the last inequality holds as mini-batch size $b$ is smaller than dataset size $n$.

Similar to the previous subsection, we derive upper an upper bound on $\sum_{t=1}^{T}\E[\|p_t\|^2]$. By recursion of $p_{t+1}$'s, we have:
$$\begin{aligned}
\mathbb{E}\|p_{t+1}\|^2 &= \mathbb{E}\|(1-Dh)p_t - h\tilde{\nabla}_t + \sqrt{2Dh}\xi_t\|^2\\
&= \mathbb{E}\|(1-Dh)p_t - h\nabla f(\theta_t) - h\Delta_t + \sqrt{2Dh}\xi_t\|^2\\
&= \mathbb{E}\|(1-Dh)p_t - h\nabla f(\theta_t)\|^2 + h^2\mathbb{E}[\|\Delta_t\|^2] + 2Dhd\\
&\leq (1-Dh)^2\mathbb{E}\|p_t\|^2 + 2Gnh(1-Dh)\sqrt{\mathbb{E}\|p_t\|^2} + h^2n^2G^2 \\
&\qquad + h^2\mathbb{E}[\|\Delta_t\|^2] + 2Dhd
\end{aligned}$$
The third equality holds because $\E[\Delta_t] = \E[\xi_t] = 0$ and $\Delta_t$ and $\xi_t$ are independent. The first inequality takes advantage of $\|\nabla f\|\leq nG$ and $\E\|p_t\|\leq \sqrt{\E\|p_t\|^2}$.

Define $S = \sum_{t=1}^T\mathbb{E}[\|p_t\|^2]$. Then, taking a grand summation over $t = 0, 1, \ldots, T-1$,
$$\begin{aligned}
S&\leq (1-Dh)^2 S + 2nGh(1-Dh)\sum_{t=0}^{T-1}\sqrt{\mathbb{E}\|p_t\|^2}\\
&\qquad + T(h^2n^2G^2 + 2Dhd) +  h^2\sum_{t=0}^{T-1}\mathbb{E}[\|\Delta_t\|^2]\\
&\leq (1-Dh)^2 S + 2nGh(1-Dh)\sqrt{T} \sqrt{S}\\
&\qquad + T(h^2n^2G^2 + 2Dhd) + \frac{8h^4L^2n^4}{b^3}S
\end{aligned}$$

Rearranging the terms we have:
$$\begin{aligned}
&(1-(1-Dh)^2 - \frac{8h^4L^2n^4}{b^3}) \frac{S}{T} + 2nGh(1-Dh)\sqrt{\frac{S}{T}} + (h^2n^2G^2 + 2Dhd) \leq 0
\end{aligned}$$

Solving a quadratic equation with respect to $\sqrt{S / T}$ and ignoring constant factors, we have:
$$\sqrt{\frac{S}{T}} \lesssim \frac{nG}{D-h^3L^2n^4b^{-3}} + \frac{\sqrt{n^2G^2 + D^2d}}{D-h^3L^2n^4b^{-3}} \lesssim \frac{\sqrt{n^2G^2 + D^2d}}{D-h^3L^2n^4b^{-3}}$$

Plugging it in
$$\sum_{t=0}^{T-1} \mathbb{E}[\|\Delta_t\|^2]\leq \frac{8h^2L^2 n^4 }{b^3}\sum_{t=0}^{T-1}\mathbb{E}\|p_t\|^2$$
we have:
\begin{equation}\label{eq:sagapart1}
\frac{1}{T}\sum_{t=1}^T \mathbb{E}[\|\Delta_t\|^2]\\
\lesssim \frac{h^2L^2 n^4 }{b^3}\left(\frac{\sqrt{n^2G^2 + D^2d}}{D-h^3L^2n^4b^{-3}}\right)^2
\end{equation}

Similar to (\ref{eq:part2}), we can bound (\ref{eq:sagafirst}) as follows:
\begin{align}
& \sum_{t=0}^{T-1} \E[\|\Delta_t\|^2] \notag \\
&\leq \frac{n^2}{b}\sum_{t=0}^{T-1}\E_{i\in [n]}[\|\nabla f_i(\theta_t) - \nabla f_i(\alpha_t^i)\|^2]  \notag \\
&\leq \frac{4Tn^2G^2}{b} \label{eq:sagapart2}
\end{align}
where (\ref{eq:sagapart2}) holds due to all $f_i$'s are $G$-Lipschitz and Cauchy's inequality $\|a+b\|^2\le 2(\|a\|^2+\|b\|^2)$.

Now, the proof of Theorem \ref{vrhmc-saga} is finished by combining (\ref{eq:corr}), (\ref{eq:sagapart1}) and (\ref{eq:sagapart2}).

\qed

\subsection{Proof of Theorem \ref{svrg2nd}}
To prove this theorem, we need the following theorem from \citep{chen2015convergence}.
The theorem shows that SGHMC with symmetric splitting can improve the dependency of MSE on step size $h$, thus allowing larger step size and faster MSE convergence.
\begin{theorem}[\citep{chen2015convergence}]
Let $\tilde{\nabla}_t$ be an unbiased estimate of $\nabla f(\theta_t)$ for all $t$. Then under Assumption \ref{3rd}, for a smooth test function $\phi$, the MSE of SGHMC with symmetric splitting is bounded in the following way:
\begin{equation}\label{eq:thm3rd}
\E(\hat{\phi} - \bar{\phi})^2\lesssim \frac{\frac{1}{T}\sum_{t=0}^{T-1}\E(\Delta V_t \psi(\theta_t, p_t))^2}{T} + \frac{1}{Th} + h^4
\end{equation}
\end{theorem}

Now, we define $\Delta_t = \tilde{\nabla}_t - \nabla f(\theta_t + \frac{h}{2}p_t)$. According to Assumption \ref{strong}, we have:
\begin{equation}\label{eq:corr2nd}
\E(\hat{\phi} - \bar{\phi})^2
\leq \frac{1}{T^2}\sum_{t=0}^{T-1}\E\|\Delta_t\|^2+ \frac{1}{Th} + h^4
\end{equation}

According to (\ref{eq:corr2nd}), we mainly need to bound the term $\sum_{t=0}^{T-1}\E\|\Delta_t\|^2$ for our SVRG2nd-HMC algorithm.
First, we unfold the definition of $\tilde{\nabla}_t$,
\begin{align*}
\tilde{\nabla}_t &= -\nabla\log \Pr(\theta_t + \frac{h}{2}p_t) + \frac{n}{b}\sum_{i\in I}(\nabla f_i(\theta_t + \frac{h}{2}p_t) - \nabla f_i(\theta_{\lfloor\frac{t}{K}\rfloor K} + \frac{h}{2}p_{\lfloor\frac{t}{K}\rfloor K})) \\
&\quad \qquad + \sum_{i=1}^n \nabla f_i(\theta_{\lfloor\frac{t}{K}\rfloor K} + \frac{h}{2}p_{\lfloor\frac{t}{K}\rfloor K})
\end{align*}

Then,
\begin{align}
\E\|\Delta_t\|^2 &= \E\|\frac{n}{b}\sum_{i\in I}(\nabla f_i(\theta_t + \frac{h}{2}p_t) - \nabla f_i(\theta_{\lfloor\frac{t}{K}\rfloor K} + \frac{h}{2}p_{\lfloor\frac{t}{K}\rfloor K})) \notag\\
&\quad \qquad - \sum_{i=1}^n(\nabla f_i(\theta_t + \frac{h}{2}p_t) - \nabla f_i(\theta_{\lfloor\frac{t}{K}\rfloor K} + \frac{h}{2}p_{\lfloor\frac{t}{K}\rfloor K}))\|^2 \notag\\
&\leq \E\|\frac{n}{b}\sum_{i\in I}(\nabla f_i(\theta_t + \frac{h}{2}p_t) - \nabla f_i(\theta_{\lfloor\frac{t}{K}\rfloor K} + \frac{h}{2}p_{\lfloor\frac{t}{K}\rfloor K}))\|^2 \notag\\
&\leq \frac{n^2}{b}\E_{i\in [n]}\|\nabla f_i(\theta_t + \frac{h}{2}p_t) - \nabla f_i(\theta_{\lfloor\frac{t}{K}\rfloor K} + \frac{h}{2}p_{\lfloor\frac{t}{K}\rfloor K})\|^2 \label{eq:first2nd}\\
&\leq \frac{n^2L^2}{b} \E\|\theta_t + \frac{h}{2}p_t - \theta_{\lfloor\frac{t}{K}\rfloor K} - \frac{h}{2}p_{\lfloor\frac{t}{K}\rfloor K}\|^2 \notag\\
&\leq \frac{n^2L^2K}{b}\sum_{j=\lfloor\frac{t}{K}\rfloor K}^{t-1} \E\|\theta_{j+1} + \frac{h}{2}p_{j+1} - \theta_j - \frac{h}{2}p_j\|^2 \notag
\end{align}

Taking a summation we have:
$$\begin{aligned}
\sum_{t=0}^{T-1}\E\|\Delta_t\|^2 &\leq \frac{n^2L^2K}{b}\sum_{t=0}^{T-1}\sum_{j=\lfloor\frac{t}{K}\rfloor K}^{t-1} \E\|\theta_{j+1} + \frac{h}{2}p_{j+1} - \theta_j - \frac{h}{2}p_j\|^2\\
&\leq \frac{n^2L^2K^2}{b}\sum_{t=0}^{T-1}\E\|\theta_{t+1} + \frac{h}{2}p_{t+1} - \theta_t - \frac{h}{2}p_t\|^2\\
&= \frac{n^2L^2K^2h^2}{b}\sum_{t=0}^{T-1}\E\|p_{t+1}\|^2
\end{aligned}$$

The last equality follows by the recursion $\theta_{t+1} = \theta_t + \frac{h}{2}p_{t+1} + \frac{h}{2}p_t$.

By definition of $p_{t+1}$ we have:
$$\begin{aligned}
&\E \|p_{t+1}\|^2 \\
=& \E \|e^{-Dh/2}\big(e^{-Dh/2}p_t - h\tilde{\nabla}_t + \sqrt{2Dh}\xi_t\big)\|^2\\
=& e^{-Dh} \E\|e^{-Dh/2}p_t - h\Delta_t - h\nabla f(\theta_t + \frac{h}{2}p_t) + \sqrt{2Dh}\xi_t\|^2\\
=& e^{-Dh} \Big(\E\|e^{-Dh/2} p_t - h\nabla f(\theta_t + \frac{h}{2}p_t)\|^2 + \E\|\sqrt{2Dh}\xi_t\|^2 + \E\|h\Delta_t\|^2\Big)\\
=& e^{-Dh} \Big(e^{-Dh}\E\|p_t\|^2 + 2e^{-Dh/2}nG\E\|p_t\| + n^2G^2h^2 + 2Dhd + h^2\E\|\Delta_t\|^2\Big)\\
\leq& (1-\frac{Dh}{4})^2 \Big((1-\frac{Dh}{4})^2\E\|p_t\|^2 + 2(1-\frac{Dh}{4})nG\sqrt{\E\|p_t\|^2} + n^2G^2h^2 + 2Dhd + h^2\E\|\Delta_t\|^2\Big)\\
\end{aligned}$$

Define $S = \sum_{t=1}^T \E\|p_t\|^2$ and $M=n^2G^2h^2 + 2Dhd$. Taking a grand summation of the above inequality for $t = 0, 1, 2, \ldots, T-1$, we have:
$$\begin{aligned}
S &\leq (1 - \frac{Dh}{4})^2 \bigg( (1 - \frac{Dh}{4})^2S + 2(1 - \frac{Dh}{4})nG\sum_{t=0}^{T-1}\sqrt{\E\|p_t\|^2} +TM +h^2\sum_{t=0}^{T-1}\E\|\Delta_t\|^2\bigg)\\
&\leq (1 - \frac{Dh}{4})^2 \bigg( (1 - \frac{Dh}{4})^2S + 2(1 - \frac{Dh}{4})nG\sqrt{TS}+TM +h^2\sum_{t=0}^{T-1}\E\|\Delta_t\|^2 \bigg)\\
&\leq (1 - \frac{Dh}{4})^2 \bigg( (1 - \frac{Dh}{4})^2S + 2(1 - \frac{Dh}{4})nG\sqrt{TS}+ TM + h^2\sum_{t=0}^{T-1}\E\|\Delta_t\|^2\bigg)\\
&\leq (1 - \frac{Dh}{4})^2 \bigg( (1 - \frac{Dh}{4})^2S + 2(1 - \frac{Dh}{4})nG\sqrt{TS}+ TM + \frac{n^2L^2K^2h^4}{b}S\bigg)
\end{aligned}$$

Rewriting it as a quadratic inequality with respect to $\sqrt{\frac{S}{T}}$, we have:
$$\Big(1 - (1 - \frac{Dh}{4})^4 - (1 - \frac{Dh}{4})^2\frac{n^2L^2K^2h^4}{b}\Big) \frac{S}{T} - 2(1 - \frac{Dh}{4})^3nG\sqrt{\frac{S}{T}} - (1-\frac{Dh}{4})^2M\leq 0$$

Solve the inequality and ignore constant factors:
\begin{align*}
\sqrt{\frac{S}{T}} &\lesssim \frac{nG + \sqrt{n^2G^2 + D^2d}}{D-L^2n^2K^2h^3b^{-1}} \\
&\lesssim \frac{\sqrt{n^2G^2 + D^2d}}{D-L^2n^2K^2h^3b^{-1}}
\end{align*}

Similar to the proof of Theorem \ref{vrhmc-svrg}, it easily follows that:
\begin{equation}\label{eq:part12nd}
\frac{1}{T}\sum_{t=0}^{T-1}\E\|\Delta_t\|^2 \lesssim \frac{L^2 n^2 K^2h^2}{b}\left(\frac{\sqrt{n^2G^2 + D^2d}}{D-L^2n^2K^2h^3b^{-1}}\right)^2
\end{equation}

Similar to (\ref{eq:part2}), we can bound (\ref{eq:first2nd}) as follows:
\begin{align}
\E[\|\Delta_t\|^2] &\leq \frac{n^2}{b}\E_{i\in [n]}\|\nabla f_i(\theta_t + \frac{h}{2}p_t) - \nabla f_i(\theta_{\lfloor\frac{t}{K}\rfloor K} + \frac{h}{2}p_{\lfloor\frac{t}{K}\rfloor K})\|^2 \notag \\
&\leq \frac{4n^2G^2}{b} \label{eq:part22nd}
\end{align}
where (\ref{eq:part22nd}) holds due to all $f_i$'s are $G$-Lipschitz and Cauchy's inequality $\|a+b\|^2\le 2(\|a\|^2+\|b\|^2)$.

Now, the proof of Theorem \ref{svrg2nd} is finished by combining (\ref{eq:corr2nd}), (\ref{eq:part12nd}) and (\ref{eq:part22nd}).
\qed

\vspace{3mm}
\subsection{Proof of Theorem \ref{saga2nd}}
Similar to the proof of Theorem \ref{svrg2nd}, we define $\Delta_t = \tilde{\nabla}_t - \nabla f(\theta_t + \frac{h}{2}p_t)$.
By Assumption \ref{strong} and Inequality (\ref{eq:thm3rd}), we have:
\begin{align*}
\E(\hat{\phi} - \bar{\phi})^2&\lesssim \frac{\frac{1}{T}\sum_{t=0}^{T-1}\E(\Delta V_t\psi(\theta_t, p_t))^2}{T} + \frac{1}{Th} + h^4 \\
&\leq \frac{1}{T^2}\sum_{t=0}^{T-1}\E\|\Delta_t\|^2+ \frac{1}{Th} + h^4
\end{align*}

Unpacking the definition of $\Delta_t$ and $\tilde{\nabla}_t$, we have:
\begin{align}
\sum_{t=0}^{T-1}\E[\|\Delta_t\|^2]
&= \sum_{t=0}^{T-1}\E[\|\tilde{\nabla}_t - \nabla f(\theta_t)\|^2] \notag\\
&= \sum_{t=0}^{T-1}\E[\|\frac{n}{b}\sum_{i\in I}\Big(\nabla f_i(\theta_t + \frac{h}{2}p_t) - \nabla f_i(\alpha_t^i)\Big) - \sum_{j=1}^n\Big(\nabla f_j(\theta_t + \frac{h}{2}p_t) - \nabla f_j(\alpha_t^j)\Big)\|^2] \notag\\
&= \sum_{t=0}^{T-1}n^2 \E[\|\frac{1}{b}\sum_{i\in I}\Big(\nabla f_i(\theta_t + \frac{h}{2}p_t) - \nabla f_i(\alpha_t^i)\Big) - \frac{1}{n}\sum_{j=1}^n\Big(\nabla f_j(\theta_t + \frac{h}{2}p_t) - \nabla f_j(\alpha_t^j)\Big)\|^2] \notag\\
&\leq \frac{n^2}{b}\sum_{t=0}^{T-1}\E_{i\in [n]}[\|\nabla f_i(\theta_t + \frac{h}{2}p_t) - \nabla f_i(\alpha_t^i)\|^2] \label{eq:sagafirst2nd}\\
&\leq \frac{L^2 n^2}{b}\sum_{t=0}^{T-1}\E_{i\in [n]}[\|\theta_t + \frac{h}{2}p_t - \alpha_t^i\|^2] \label{eq:saga2ndtmp}
\end{align}

Let $\gamma = 1 - (1 - 1/n)^b$. Then,
$$\begin{aligned}
&\E[\|\theta_t  + \frac{h}{2}p_t - \alpha_t^i\|^2] \\
&= \sum_{j=0}^{t-1}\E[\|\theta_t + \frac{h}{2}p_t  - \theta_j  - \frac{h}{2}p_j \|^2] \Pr(\alpha_t^i = \theta_j + \frac{h}{2}p_j)\\
&= \sum_{j=0}^{t-1}\E[\|\theta_t + \frac{h}{2}p_t  - \theta_j  - \frac{h}{2}p_j \|^2] (1-\gamma)^{t-j-1}\gamma\\
&= h^2\sum_{j=0}^{t-1} \E[\|p_t + p_{t-1} + \cdots + p_{j+1}\|^2] (1-\gamma)^{t-j-1}\gamma\\
&\leq h^2\sum_{j=0}^{t-1} (t-j) (1-\gamma)^{t-j-1}\gamma (\E[\|p_t\|^2] + \E[\|p_{t-1}\|^2] + \cdots + \E[\|p_{j+1}\|^2])\\
&= h^2\sum_{j=1}^{t}\E[\|p_j\|^2] \sum_{k=0}^{j-1}(t-k)(1-\gamma)^{t-k-1}\gamma\\
&\leq h^2\sum_{j=1}^{t}\E[\|p_j\|^2] \sum_{k=0}^\infty (t-j+k+1)(1-\gamma)^{t-j+k}\gamma\\
&< h^2\sum_{j=1}^t \E[\|p_j\|^2] (\frac{1}{\gamma} + t - j) (1-\gamma)^{t-j}
\end{aligned}$$

Summing over all $t$ and $i$, we then have:
\begin{align}
\sum_{t=0}^{T-1}\E_{i\in[n]}[\|\theta_t + \frac{h}{2}p_t - \alpha_t^i\|^2]
&\leq h^2\sum_{t=0}^{T-1}\sum_{j=1}^t \E[\|p_j\|^2] (\frac{1}{\gamma} + t - j) (1-\gamma)^{t-j}\notag\\
&= h^2\sum_{t=0}^{T-1}\sum_{j=1}^t \E[\|p_j\|^2] (\frac{1}{\gamma} + t - j) (1-\gamma)^{t-j}\notag\\
&\leq h^2\sum_{t=1}^{T-1} \E[\|p_t\|^2]\sum_{j=t}^{T-1} (\frac{1}{\gamma} + j-t)(1-\gamma)^{j-t}\notag\\
&\leq h^2\sum_{t=1}^{T-1} \E[\|p_t\|^2]\sum_{j=t}^\infty (\frac{1}{\gamma} + j-t)(1-\gamma)^{j-t}\notag\\
&\leq h^2\sum_{t=1}^{T-1} \E[\|p_t\|^2]\frac{2}{\gamma^2} = \frac{2}{\gamma^2}h^2\sum_{t=1}^{T-1} \E[\|p_t\|^2]\notag\\
&\leq \frac{8h^2n^2}{b^2}\sum_{t=1}^{T} \E[\|p_t\|^2] \label{eq:saga2ndtmp2}
\end{align}

The last inequality is because $(1-1/n)^b \leq \frac{1}{1 + \frac{b}{n-1}}$, and thus $\gamma = 1 - (1-1/n)^n \geq \frac{\frac{b}{n-1}}{1 + \frac{b}{n-1}} > \frac{b}{2n}$; the last inequality holds as mini-batch size $b$ is smaller than dataset size $n$.

Now, we derive an upper bound on $\sum_{t=1}^{T}\E[\|p_t\|^2]$. By recursion of $p_{t+1}$'s, we have:
$$\begin{aligned}
&\E \|p_{t+1}\|^2 \\
&= \E \|e^{-Dh/2} (e^{-Dh/2}p_t - h\tilde{\nabla}_t + \sqrt{2Dh}\xi_t)\|^2\\
&= e^{-Dh} \E\|e^{-Dh/2}p_t - h\Delta_t - h\nabla f(\theta_t + \frac{h}{2}p_t) + \sqrt{2Dh}\xi_t\|^2\\
&= e^{-Dh} \Big(\E\|e^{-Dh/2} p_t - h\nabla f(\theta_t + \frac{h}{2}p_t)\|^2 + \E\|\sqrt{2Dh}\xi_t\|^2 + \E\|h\Delta_t\|^2\Big)\\
&= e^{-Dh} \Big(e^{-Dh}\E\|p_t\|^2 + 2e^{-Dh/2}nG\E\|p_t\|  + n^2G^2h^2 + 2Dhd + h^2\E\|\Delta_t\|^2\Big)\\
&\leq (1-\frac{Dh}{4})^2 \Big((1-\frac{Dh}{4})^2\E\|p_t\|^2 + 2(1-\frac{Dh}{4})nG\sqrt{\E\|p_t\|^2}  + n^2G^2h^2 + 2Dhd + h^2\E\|\Delta_t\|^2\Big)
\end{aligned}$$

Define $S = \sum_{t=1}^T \E\|p_t\|^2$ and $M=n^2G^2h^2 + 2Dhd$. Taking a grand summation of the above inequality for $t = 0, 1, 2, \ldots, T-1$, we have:
$$\begin{aligned}
S &\leq (1 - \frac{Dh}{4})^2 \bigg( (1 - \frac{Dh}{4})^2S + 2(1 - \frac{Dh}{4})nG\sum_{t=0}^{T-1}\sqrt{\E\|p_t\|^2} + TM + h^2\sum_{t=0}^{T-1}\E\|\Delta_t\|^2\bigg)\\
&\leq (1 - \frac{Dh}{4})^2 \left( (1 - \frac{Dh}{4})^2S + 2(1 - \frac{Dh}{4})nG\sqrt{TS}+ TM + h^2\sum_{t=0}^{T-1}\E\|\Delta_t\|^2\right)\\
&\leq (1 - \frac{Dh}{4})^2 \left( (1 - \frac{Dh}{4})^2S + 2(1 - \frac{Dh}{4})nG\sqrt{TS}+ TM + h^2\sum_{t=0}^{T-1}\E\|\Delta_t\|^2\right)\\
&\leq (1 - \frac{Dh}{4})^2 \left( (1 - \frac{Dh}{4})^2S + 2(1 - \frac{Dh}{4})nG\sqrt{TS}+ TM + \frac{8n^4L^2h^4}{b^3}S\right)
\end{aligned}$$

Similar to the proof of theorem \ref{svrg2nd}, we solve a quadratic inequality with respect to $\sqrt{\frac{S}{T}}$, and then,
\begin{equation}\label{eq:saga2ndtmp3}
\sqrt{\frac{S}{T}}\lesssim \frac{\sqrt{n^2G^2 + D^2d}}{D-h^3L^2n^4b^{-3}}
\end{equation}

From (\ref{eq:saga2ndtmp}), (\ref{eq:saga2ndtmp2}), (\ref{eq:saga2ndtmp3}) and the definition of $S$, we have:
\begin{align}
&\sum_{t=0}^{T-1} \mathbb{E}[\|\Delta_t\|^2] \notag\\
&\leq \frac{8h^2L^2 n^4 }{b^3}\sum_{t=0}^{T-1}\mathbb{E}\|p_t\|^2 \notag\\
&\leq \frac{8h^2L^2 n^4T}{b^3}\left(\frac{\sqrt{n^2G^2 + D^2d}}{D-h^3L^2n^4b^{-3}}\right)^2 \label{eq:sagapart12nd}
\end{align}

Similar to (\ref{eq:part2}), we can bound (\ref{eq:sagafirst2nd}) as follows:
\begin{align}
& \sum_{t=0}^{T-1} \E[\|\Delta_t\|^2] \notag \\
&\leq \frac{n^2}{b}\sum_{t=0}^{T-1}\E_{i\in [n]}[\|\nabla f_i(\theta_t + \frac{h}{2}p_t) - \nabla f_i(\alpha_t^i)\|^2]  \notag \\
&\leq \frac{4Tn^2G^2}{b} \label{eq:sagapart22nd}
\end{align}
where (\ref{eq:sagapart22nd}) holds due to all $f_i$'s are $G$-Lipschitz and Cauchy's inequality $\|a+b\|^2\le 2(\|a\|^2+\|b\|^2)$.

Now, the proof of Theorem \ref{saga2nd} is finished by combining (\ref{eq:corr2nd}), (\ref{eq:sagapart12nd}) and (\ref{eq:sagapart22nd}).
\qed

\end{document}